\documentclass[11pt, letterpaper, logo, onecolumn, copyright]{main}

\usepackage[authoryear, sort&compress, round]{natbib}

\usepackage[inkscapeformat=png]{svg}

\usepackage[most, breakable, skins]{tcolorbox}

\tcbuselibrary{skins}
\usepackage{lipsum}
\usepackage{tabularx}
\usepackage{afterpage}
\usepackage{booktabs}
\usepackage{subcaption}
\usepackage{makecell}
\usepackage{multirow}
\usepackage{bm}
\usepackage{multicol}
\usepackage{array}
\usepackage{float}
\usepackage{listings, listings-rust}
\usepackage{fontawesome5}
\usepackage{hyperref}
\usepackage{amssymb,graphicx}
\usepackage[dvipsnames]{xcolor}
\usepackage{cleveref}
\usepackage{longtable}
\usepackage{pdflscape}
\usepackage{adjustbox}
\usepackage{nicematrix}
\usepackage{CJKutf8}
\usepackage{ragged2e}
\usepackage{colortbl}
\usepackage{enumitem}
\usepackage[ruled,linesnumbered]{algorithm2e}
\usepackage{pifont}
\usepackage{amsmath}
\usepackage{amsthm}  
\usepackage{circledsteps}
\usepackage{diagbox}
\usepackage{bookmark}

\usepackage{wrapfig}
\usepackage{xspace}
\usepackage{tikz}
\usepackage[normalem]{ulem}

\usepackage{pgfplots}
\usepackage{pgfplotstable}
\pgfplotsset{compat=1.18}

\definecolor{medgray55}{gray}{0.55}
\definecolor{medgray}{gray}{0.7}
\definecolor{litegray}{gray}{0.9}
\definecolor{gblue}{RGB}{210, 227, 252}
\definecolor{gred}{RGB}{250, 210, 207}
\definecolor{gyellow}{RGB}{254, 239, 195}
\definecolor{ggreen}{RGB}{206, 234, 214}
\definecolor{gorange}{RGB}{254, 223, 200}

\definecolor{gblue9}{RGB}{23, 78, 166}
\definecolor{gred9}{RGB}{165, 14, 14}
\definecolor{gyellow9}{RGB}{227, 116, 0}
\definecolor{ggreen9}{RGB}{13, 101, 45}
\definecolor{gorange9}{RGB}{176, 96, 0}

\definecolor{myblue}{rgb}{0,0,1}
\definecolor{myred}{rgb}{1,0,0}
\definecolor{mylightgray}{gray}{0.95}
\definecolor{myCite}{HTML}{1C4587}

\definecolor{highlightblue}{HTML}{185ABC}
\definecolor{cellHighlight}{HTML}{dbefff}

\usepackage{minitoc}

\noptcrule


\newcolumntype{L}[1]{>{\raggedright\let\newline\\\arraybackslash\hspace{0pt}}m{#1}}
\newcolumntype{C}[1]{>{\centering}m{#1}}

\newcolumntype{R}[1]{>{\raggedleft\let\newline\\\arraybackslash\hspace{0pt}}m{#1}}

\definecolor{ao}{rgb}{0.0, 0.0, 1.0}

\newcommand\vcent[1]{\vcenter{\hbox{#1}}}

\newcommand\loudspeaker[1][3]{\ensuremath{\vcent{\rule{.6ex}{.6ex}}\kern-.5ex
  \vcent{\scalebox{.6}[1]{\rotatebox[origin=center]{90}{$\blacktriangle$}}}
  \ifnum#1>0\relax\kern.05ex\vcent{\scalebox{.4}{\ttfamily)}}
  \ifnum#1>1\relax\kern-.4ex\vcent{\scalebox{.56}{\ttfamily)}}
  \ifnum#1>2\relax\kern-.55ex\vcent{\scalebox{.7}{\ttfamily)}}
  \fi\fi\fi}
}

\makeatletter
\renewcommand\subparagraph{
 \@startsection {subparagraph}{5}{\z@ }{3.25ex \@plus 1ex
 \@minus .2ex}{-1em}{\normalfont \normalsize \bfseries }}
\makeatother

\let\cite\citep
\hypersetup{
  citecolor = myCite,  
  linkcolor = myCite,   
  urlcolor  = myCite
}

\title{AgentGym2: Benchmarking Large Language Model Agents in De-Idealized Real-World Environments}
\author{
    Zhiheng Xi$^1$$^{* \dag}$,  Dingwen Yang$^1$$^*$, Jiaqi Liu$^1$, Jixuan Huang$^1$, Honglin Guo$^1$, Baodai Huang$^1$, Tinggang Chen$^1$, Qi Zhang$^1$, Zhonghang Lu$^1$, Chenyu Liu$^1$, Jiajun Sun$^1$, Jiazheng Zhang$^1$, Dingwei Zhu$^1$, Xin Guo$^1$, Junzhe Wang$^1$, Zhihao Zhang$^1$, Yuming Yang$^1$, Junjie Ye$^1$, Minghe Gao$^2$, Dongrui Liu$^3$, Jiaming Ji$^4$, Guohao Li$^5$ 
    \textbf{Tao Gui$^{1}$$^\dag$, Qi Zhang$^1$$^\dag$, Xuanjing Huang$^1$$^\dag$}
\\
$^1$Fudan University  $^2$Zhejiang University \\ $^3$Shanghaijiaotong University  $^4$Peking University $^5$CAMEL-AI.org\\
\texttt{zhxi22@m.fudan.edu.cn, \{tgui,qz,xjhuang\}@fudan.edu.cn} 
}
\vspace{-50pt}
\begin{abstract}
Language agents, i.e., LLM agents, progress rapidly and are increasingly deployed in production environments. This trend underscores the urgent need for rigorous and realistic evaluations. However, most existing benchmarks evaluate agents in simplified, idealized settings. They typically rely on pre-packaged tool interfaces, overlook critical steps, and assume inputs are clean and fully specified. Consequently, they understate the difficulty of real deployments, where uncertainty and noise are ubiquitous and agents must proactively explore the environment to uncover new tools. To bridge this gap, we present AgentGym2, a new evaluation framework with task instances grounded in real-world end-to-end working demands. Beyond reasoning and planning, it measures agents’ ability to execute end-to-end procedures, discover tools via exploration, compose tools for unseen tasks, and remain robust to noisy and underspecified information. Experiments on 15 proprietary and open-source models show that even SOTA systems like Gemini and GPT-5 struggle on AgentGym2, revealing a substantial gap between the capability of current agents and the demands of real-world applications. 
\end{abstract}

\begin{document}

\doparttoc
\faketableofcontents

\begingroup
  \renewcommand\thefootnote{}
  \footnote{\textsuperscript{*}Equal contribution.
            \textsuperscript{\dag}Corresponding authors.}
    \footnote{\textsuperscript{1}Our code and dataset are available at \url{https://github.com/hotdog-zz/Agentgym2} and \url{https://huggingface.co/datasets/hotdogzz/Agentgym2}}
  \addtocounter{footnote}{-1}
\endgroup

\vspace{-30pt}
\maketitle

\vspace{-15pt}

\begin{figure*}[h]
    \includegraphics[width=0.99\linewidth]{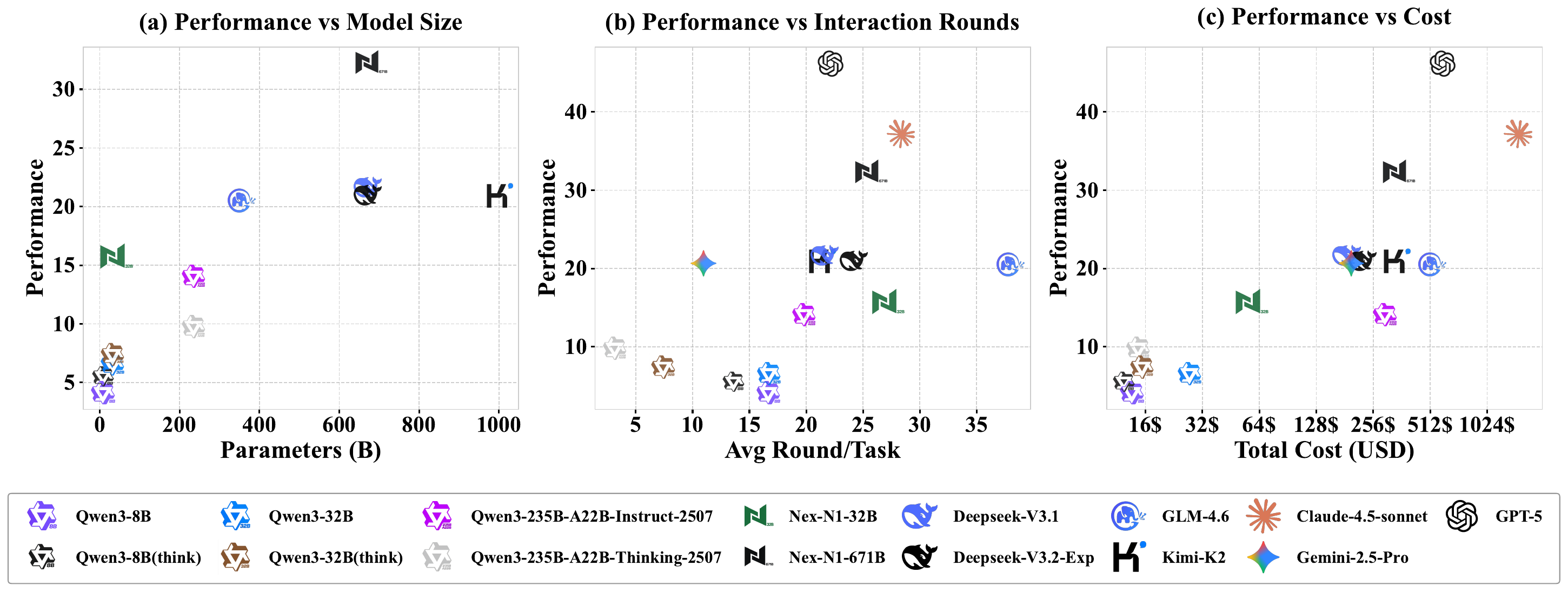}
    \centering
 	\caption{
Effect on performance across three dimensions: total cost,  the average number of interaction rounds, and the parameter scale.
  }
  \vspace{-2mm}
  \label{fig:params_performance}
\end{figure*}

\section{Introduction}

As language language models advance rapidly, their applications evolve from simple conversational chatbots to autonomous agents capable of complex tasks such as deep research and data analysis, with increasing deployment in production~\citep{li25tongyidr,zheng25deepresearcher,jin25searchr1,hong-etal-2025-data,zhu25dasurvey}. Consequently, it is essential to evaluate their ability in handling practical tasks in real-world~\citep{garg25real,patwardhan25gdpval,ko26gaia}.

Despite rapid growth in agent evaluation research, existing benchmarks remains simplified or idealized~\citep{xu24theagentcompany,patil2025bfcl}, failing to capture the complexity in real-world environment.
They typically provide pre-packaged tool sets, pre-solve intermediate yet necessary steps, and overlook the uncertainty and noise inherent in real-world scenarios. 
As a result, they may significantly underestimate real-world difficulty: user requests are often underspecified or ambiguous, information sources may be incomplete or misleading, and necessary tools may be unavailable upfront~\citep{DBLP:journals/corr/abs-2507-03336,DBLP:journals/corr/abs-2511-08798}. Consequently, several capabilities essential for reliable production deployment remain insufficiently evaluated.

To bridge this gap, we introduce AgentGym2, an evaluation framework for assessing language agents in realistic environments. It spans diverse scenarios—including data analysis, deep research and complex tool use—across over 27 domains. It emphasizes end-to-end task completion under real-world uncertainty, assessing not only standard agentic skills such as search~\citep{wei25browsecomp,zhou25browsecompzh}, planning~\citep{erdogan25lanact}, reasoning~\citep{phan25hle}, and coding~\citep{yang24sweagent,gao25traeagent}, but also an agent’s ability to proactively clarify and interpret goals, explore and discover tools through interaction, compose tools into novel workflows, carry out complete procedures, and remain robust in the presence of noise, ambiguity, and underspecified inputs.

\begin{wrapfigure}{r}{0.4\linewidth}
    \includegraphics[width=0.99\linewidth]{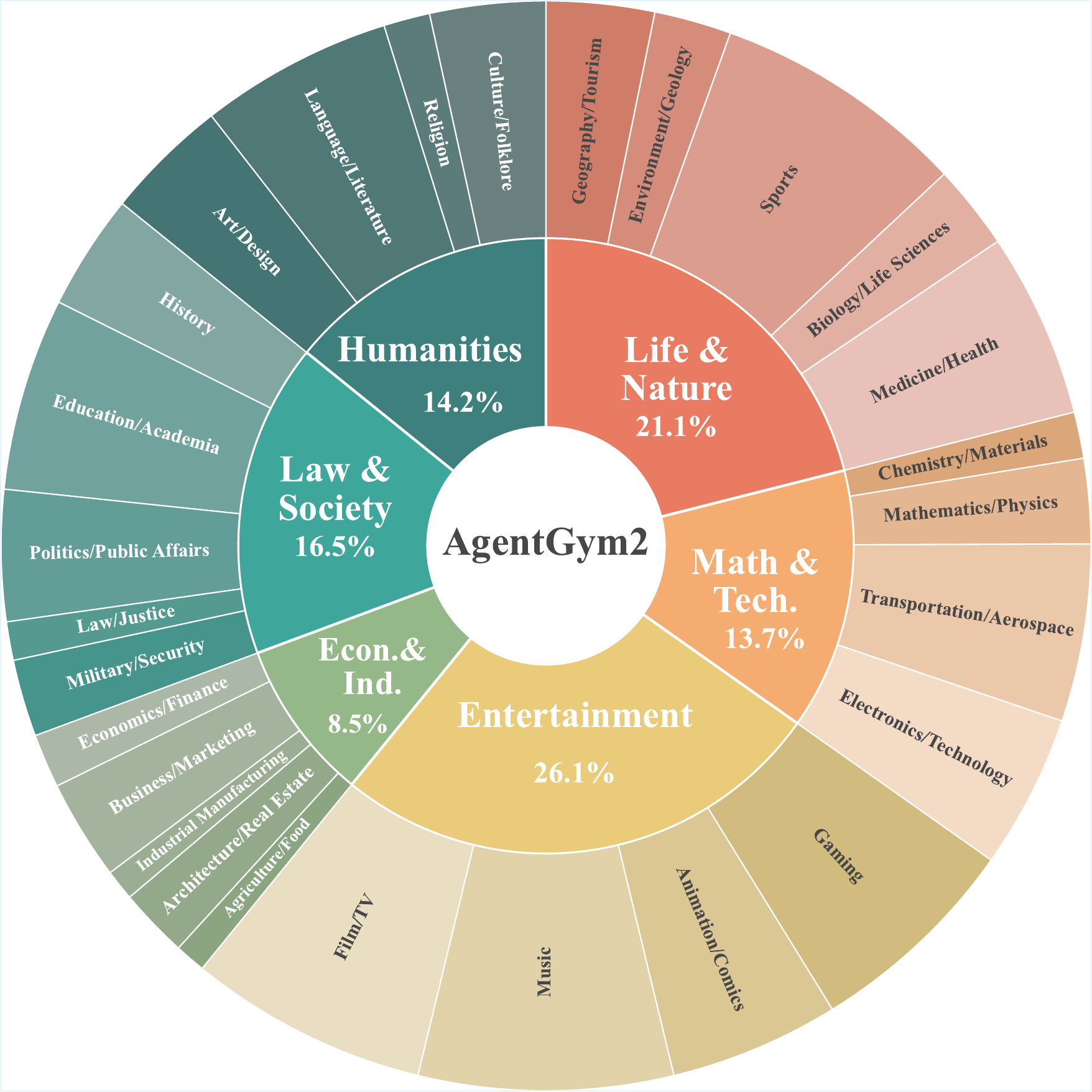}
    \centering
\caption{Domain distribution of AgentGym2. Econ. \& Ind. denotes Economy \& Industry. Math \& Tech. denotes Math \& Technology.
}
\label{fig:domain_statistic}
\vspace{-10pt}
\end{wrapfigure} 

At the system level, AgentGym2 equips agents with a basic yet composable toolbox rather than query-specific, pre-selected tools~\citep{yao22webshop,liu24agentbench,xi25agentgym}. This design exposes a large and realistic action space, encouraging agents to explore what to use and how to use it. The toolbox covers five core categories—web browsing, information retrieval, file processing, multimodal understanding, and code execution—supporting over 27 actions. Built with a modular, decoupled architecture, AgentGym2 enables environment isolation and parallel tool invocation, ensuring rigorous, reproducible, and scalable evaluation.

To ground our task instances in authentic user demands, we curate real requests from diverse platforms, including GitHub, Reddit, and Kaggle. We transform these raw needs into challenging and diverse evaluation instances through a combination of automated synthesis, query augmentation, and manual annotation. To ensure quality and consistency at scale, all tasks undergo a cascading verification pipeline that combines model-based checks with human review, validating task validity, clarity, and expected outcomes.

We evaluate 15 leading proprietary and open-source models on AgentGym2 and find that even state-of-the-art systems such as GPT-5 (\textasciitilde 44\%) and Claude Sonnet 4.5 (\textasciitilde 37\%) fail to deliver satisfactory performance, indicating that today’s agents still require substantial progress before they can be reliably deployed in production. For open-source models, strong post-training can significantly boost performance: Nex-N1-32B and Nex-N1-671B~\citep{agiteam2025nexn1agenticmodelstrained} outperform their respective base models by roughly 9.16\% and 10.53\%. We also observe consistent gains as model parameters and computational cost scale, whereas increasing the number of interaction turns does not produce a scaling effect. Fine-grained analyses in Section \ref{sec:analysis} characterize agent behaviors and reveal recurring failure patterns across scenarios, such as insufficient exploration~\citep{xi25agentgymrl} and hallucinations~\citep{sahoo-2024-hallucination}. 

In summary, our main contributions are:
\begin{itemize}
    \item We propose a new interaction framework that equips agents with a basic action space and supports further exploration, enabling isolated, parallel, and scalable interactions.
    \item We introduce a new benchmark to evaluate LLM agents under de-idealized environments and realistic requirements.
    \item We conduct extensive experiments and analyses to provide in-depth insights, aiming to support future research, development, and deployment of language agents.
\end{itemize}

\begin{table*}[t]
\centering
\resizebox{0.999\textwidth}{!}{
\begin{tabular}{lccccccc}
\toprule
\textbf{Benchmark} & \textbf{Scenario}  & \textbf{Non-Simulated Environment?} & \textbf{Realistic Action Space?} & \textbf{Tool discovery?} & \textbf{E2E Data Analysis?} & \textbf{Inserted Distractors?} \\ \midrule
BrowseComp~\citep{wei25browsecomp} & Search & yes  & yes     & no & - & no    \\
DA-Code~\citep{DBLP:conf/emnlp/HuangLYZLWHHLZL24}    & Data Analysis  & yes  & yes     & no & no     & no    \\
$\tau$-Bench~\citep{yao2025taubench}  & Tool Use  & no   & yes     & no & - & no    \\
AgentBoard~\citep{DBLP:conf/nips/MaZZYYJLKH24} & Multi-Scenarios& no   & no & no & - & no    \\
AgentBench~\citep{liu24agentbench} & Multi-Scenarios& no   & no & no & no     & no    \\
AgentGym~\citep{xi25agentgym, xi25agentgymrl} & Multi-Scenarios& no   & no & no & - & no    \\
\midrule
\rowcolor{blue!10!magenta!6}
AgentGym2 (Ours)   & Multi-Scenarios& yes  & yes     & yes     & yes    & yes   \\ \bottomrule
\end{tabular}
}
\caption{Comparison of different agent benchmarks. Specifically, Tool discovery indicates whether the necessary tools or resources are not provided at the start and must be discovered during interaction; E2E Data Analysis (incl. cleaning) indicates whether the benchmark covers an end-to-end data analysis pipeline, including data cleaning/preprocessing and analysis; and Distractors indicates whether the tasks include explicitly injected distracting information, such as noisy information in search queries or dirty/irrelevant content in data files. Our Agentgym2 is the only benchmark that roots in realistic environment and action space with all of the three dimensions.}
\label{tab:benchmark_comparison}
\end{table*}

\section{Related Works}

\paragraph{Development of language agents.}

Early large language models are primarily framed as conversational systems designed for dialogue. This paradigm shifts with the introduction of approaches such as ReAct~\citep{yao23react}, which enables language agents to iteratively interact with complex environments through action and feedback. Building on this foundation, a growing body of work develops domain-specialized agents spanning software engineering~\citep{yang24sweagent,wang25openhands}, web navigation and information seeking~\citep{jin25searchr1,zheng25deepresearcher,li25tongyidr}, and data analysis pipelines~\citep{hong-etal-2025-data,guo-etal-2025-critiq,zhu25dasurvey}, highlighting the versatility of agent-based frameworks and their promise for practical deployment across a range of real-world tasks. Despite strong results on curated benchmarks, existing agent architectures rely on idealized, closed-world assumptions with pre-specified tools, limiting their effectiveness in unconstrained, real-world workflows.

\paragraph{Benchmarks for evaluating language agents.}

Recent benchmarks have begun to evaluate language agents in richly simulated environments that span a wide range of interactive tasks, including web navigation and shopping~\citep{yao22webshop}, embodied and household tasks~\citep{shridhar21alfworld}, scientific discovery environments~\citep{wang-etal-2022-scienceworld}, text-based games~\citep{prasad24textcraft}, and tool-using scenarios~\citep{patil2025bfcl,yao2025taubench,barres25tau2}. By integrating these heterogeneous tasks under a unified interaction interface and evaluation protocol, frameworks such as AgentBench~\citep{liu24agentbench} and AgentGym~\citep{xi25agentgym} provide a standardized testbed for assessing agent behavior across diverse environments. Beyond such wide-spectrum platforms, the community has also developed a growing number of domain-specific benchmarks tailored to real-world application areas, most notably software engineering~\citep{yang25swe,deng25swepro,zan25multiswe,tbench_2025}, data science workflows~\citep{lai23ds1000,lei2025dacomp,wang25fdabench,egg25dabstep} and composing multiple well-formatted real MCP tools~\citep{wang2025mcpbench,mo2025livemcpbench,guo2025mcpagentbench}. Despite these advances, existing environments often rely on sandboxed settings with curated tools and well-specified instructions, which limits their ability to assess open-ended tool discovery and robustness. We address this limitation with a non-idealized evaluation suite focused on end-to-end execution and dynamic tool acquisition. We provide a comparison on core dimensions in real deployment settings in Table~\ref{tab:benchmark_comparison}.

\begin{figure*}[t]
\includegraphics[width=0.99\linewidth]{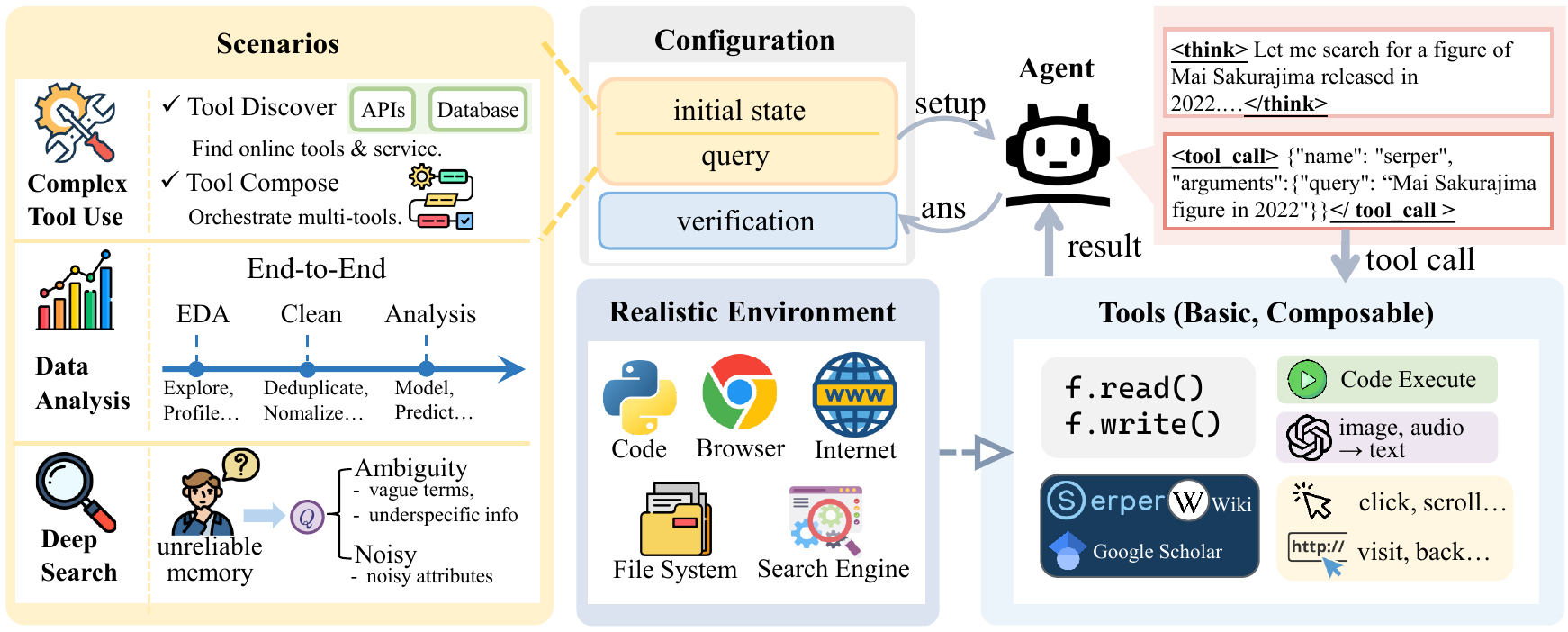}
\centering
\vspace{-1pt}
\caption{
Overview of the AgentGym2 framework. It features a layered and modular design that decouples agents from tools and environments, comprising the following components: Environment, Tools, Configurations, and Agent. It includes there cenarios, and the tasks are grounded in non-idealized, complex real-world conditions.
}
\vspace{-1mm}
\label{fig:main}
\end{figure*}

\section{AgentGym2 Framework}
\subsection{Architecture Overview}
AgentGym2 adopts a layered and modular design that decouples agents from tools and environments~\citep{li2023camel,xi25agentgym,ko26gaia}. As illustrated in Figure \ref{fig:main}, the architecture comprises the following components:
\begin{itemize}
    \item \textbf{Environment}: The realistic digital runtime that provides underlying infrastructure and services for agents to execute tasks.
    \item \textbf{Tools}: The interfaces exposed by environments. In AgentGym2, these are general-purpose, basic, and composable tools.
    \item \textbf{Agent}: The LLM agent that receives observations from environments via the interfaces, performs reasoning over its interaction trajectory, and outputs actions. Note that an agent may invoke multiple tools simultaneously within a single step.
    \item \textbf{Configurations}: Composed of a user request, a specified initial state, and the corresponding task verification mechanism. Inputs for agents encompass multiple modalities, including text and images, and a single scenario may involve multiple environments.
\end{itemize}
\subsection{Features and Characteristics}
AgentGym2 extends AgentGym~\citep{xi25agentgym,xi25agentgymrl}, which provides \textbf{simulated} environments and corresponding \textbf{specialized} interfaces for LLM agents. In this work, we extend it with realistic digital environments~\citep{ko26gaia}, construct a basic, general-purpose, and composable toolbox, and implement sophisticated engineering designs to support scalable evaluation in real-world settings.

\paragraph{Realistic scenarios and environments.}
We first identify three high-demand real-world scenarios for agentic tasks: complex tool usage, data analysis, and deep search. 
These demand high-level reasoning and the ability to handle heterogeneous data. AgentGym2 supports these requirements by providing a reliable infrastructure across five domains: code execution, file systems, search engines, web browsers and internet. We then synthesize user requests grounded in these real-world task distributions, requiring agents to interact with multimodal inputs like webpage snapshots and source code. This approach ensures our benchmark reflects practical challenges rather than environment-specific limitations~\citep{liu24agentbench,xi25agentgym}.
\paragraph{Basic and composable toolbox.}
Instead of pre-selecting domain-specific tools, AgentGym2 provides a basic, general-purpose, and composable toolbox covering five core categories: web browsing, information retrieval, file processing, multimodal understanding, and code execution. We achieve this by collecting a suite of real-world tools, fixing their existing bugs, and extending them to support isolated runtime, ultimately integrating them into a unified toolbox. Notably, we grant agents full access to the entire internet rather than limiting them to pre-restricted websites, enabling them to actively discover and learn the usage of additional tools as needed. This design provides agents with a large and realistic action space.
\paragraph{Reliable engineering design.}
To ensure scalable, precise, and reproducible evaluation, AgentGym2 implements sophisticated engineering optimizations.  First, it ensures strict runtime isolation to prevent erroneous evaluations, including file management systems to avoid read-write conflicts, isolated browser instances to prevent history-based cheating, and sandboxed environments for safe code execution~\citep{yang25swe,pan25swegym}. Moreover, AgentGym2 supports parallel tool invocation via asynchronous execution, enabling agents to initiate multiple tool calls concurrently within a single interaction. This design allows simultaneous querying of multiple sources or parallel execution of tasks, ensuring efficient resource use.

\begin{figure*}[t]
\centering
\includegraphics[width=0.999\linewidth]{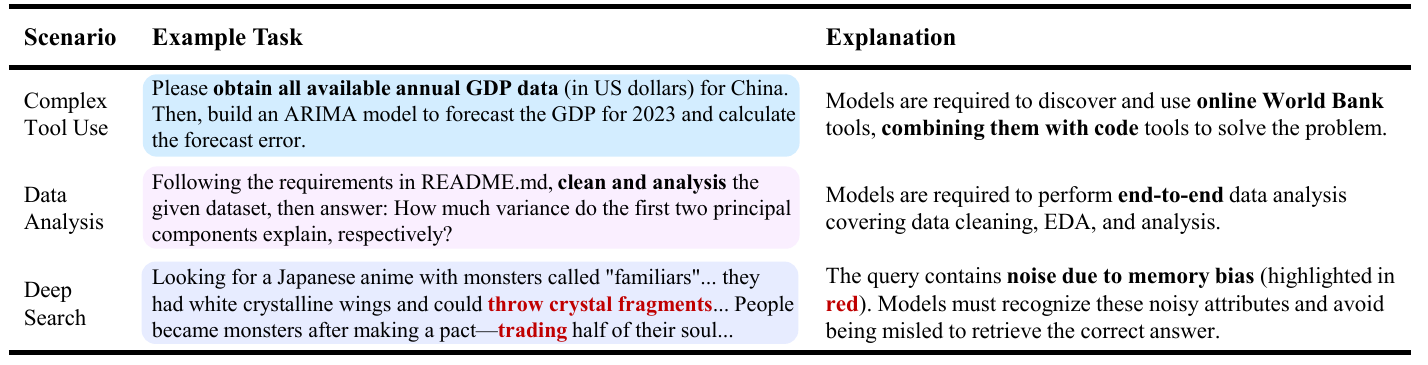}
\vspace{-18pt}
\caption{Examples of three distinct scenarios in AgentGym2.
For complex tool using scenario, we focuses on the capability of LLM on tool discovery and composition.
For data analysis, AgentGym2 challenges agents’ end-to-end data analysis
capabilities, including EDA procedures, data processing, and in-depth analysis.
To simulate the noise in the real-world user, we consider the ambiguity and bias in deep search scenario.
} 
\vspace{-5pt}
\label{fig:task_examples}
\end{figure*}

\section{Benchmark Construction}
Besides basic agentic skills such as reasoning, planning, and coding, AgentGym2 targets a broader set of capabilities: \textbf{proactively clarifying and interpreting goals, discovering tools through interaction, composing tools into novel workflows, executing complete procedures end to end, and remaining robust to noise, ambiguity, and underspecified inputs}. 

We next describe our data construction methodology. In total, AgentGym2 covers 437 tasks spanning 27 domains, the distribution is in Figure~\ref{fig:domain_statistic}. The key statistics is in Table~\ref{tab:statistics}, and the cases of three scenarios can be seen in Figure~\ref{fig:task_examples}.

\begin{wraptable}{r}{8.5cm}
\centering
\resizebox{\linewidth}{!}{
\begin{tabular}{lr}
\toprule
\textbf{Statistics}   & \textbf{Number}           \\ \midrule
Total query           & 437                       \\
- Complex Tool Use    & 182                       \\
- Data Analysis       & 57                        \\
- Deep Search         & 198                       \\
Language              & ZH/EN                     \\ \midrule
Total Attachment      & 652                       \\
Attachment Modality \ \  & Text, Image, Audio, Video \\
Basic Tool Types      & 27                        \\ \bottomrule
\end{tabular}
}
\caption{Key statistics of AgentGym2.}
\vspace{-5pt}
\label{tab:statistics}
\end{wraptable}

\paragraph{Complex tool using scenario.}
We first curate high-frequency web tools across academic, daily-life, and entertainment domains. Based on these tools, we focus on constructing two types of instances: tool discovery and tool composition, which contain 57 and 125 instances, respectively. Our instances are created through (i) manually adapting real-world discussions and questions from volunteers from diverse backgrounds, and (ii) augmenting existing benchmarks (e.g., BrowseComp~\citep{wei25browsecomp}) by having our authors replace key query entities with external attachments, such as CSV/JSON files or manually curated images, which may include distractors. A rigorous, iterative verification process ensures that each task is solvable using the provided basic tools—potentially by discovering additional interfaces or composing novel workflows—and that reaching the solution requires processing the attachments and yields a unique, verifiable answer.

\paragraph{Data analysis scenario.}
Previously, data analysis benchmarks typically provide preprocessed, high-quality data and focused only on final outcomes, while overlooking necessary steps such as Exploratory Data Analysis (EDA) and data cleaning~\citep{lai23ds1000,wang25fdabench,egg25dabstep}. As a result, the evaluation conditions are inconsistent with real-world production settings. Hence, our authors construct 57 instances based on sources such as GitHub and Kaggle by enriching real-world problems, injecting target challenges, and manually building tasks from scratch. These tasks challenge agents’ end-to-end data analysis capabilities, including EDA procedures, data processing, and in-depth analysis. A quality control step similar to the one used in complex tool use was also conducted to ensure data quality.

\paragraph{Deep search scenario.}
Deep Search requires iterative retrieval, synthesis, denoising, and analysis across multiple sources ~\citep{wei25browsecomp,zhou25browsecompzh,jin25searchr1}. Real queries can be well-specified, but often include ambiguity or memory-induced inaccuracies—cases underrepresented in prior benchmarks. We therefore generate two query types via controlled perturbations: (i) Ambiguity-only (114 instances) with entity/relational ambiguity; and (ii) Ambiguity-with-bias (84 instances) combining ambiguity with inaccurate attributes. This setting evaluates both core deep-search skills (retrieval, multi-hop reasoning, cross-source synthesis) and robustness to underspecified, irrelevant, and noisy information. Ambiguity-only instances are automatically synthesized with quality control by selecting a Wikipedia answer entity, recursively sampling relations to introduce ambiguity, merging them into multi-hop questions, and verifying difficulty and answer uniqueness. Ambiguity-with-bias instances are sourced from memory-based social-media queries (e.g., Reddit, Xiaohongshu) or are augmented by injecting noisy attributes with hedging phrases to ambiguity-only instances; all instances are manually checked to ensure the unique answer remains unchanged.

\paragraph{Answer verify.}
All reference answers in AgentGym2 consist of short strings or numbers. This format enables straightforward evaluation through LLM-based judges that assess semantic equivalence between predicted and reference answers.

\section{Experiments}

\subsection{Experimental Setups}
\paragraph{Models.}
We evaluate both proprietary and open-source models. Proprietary models include GPT-5, Claude-4.5-Sonnet, and Gemini-2.5-Pro~\citep{gemini25}. Open-source models include Kimi-K2~\citep{bai25k2}, Qwen3 series (Qwen3-8B, Qwen3-32B, Qwen3-235B-A22B-2507)~\citep{yang25qwen3}, Nex-N1 series (Nex-N1-32B, Nex-N1-671B)~\citep{agiteam2025nexn1agenticmodelstrained}, GLM-4.6~\citep{zeng25glm45}, and DeepSeek series (DeepSeek-V3.1, DeepSeek-V3.2-Exp)~\citep{deepseek-v3,deepseekai2025deepseekv32pushingfrontieropen}.

\paragraph{Implementation details.}
We leverage the ReAct \citep{yao23react} style for agents, with all tools provided in the OpenAI tool schemas. We limit the number of interaction rounds to 100, and the task terminates when the agent outputs an answer enclosed within the special format "<final\_answer>...</final\_answer>". For the main results, each task is run three times with a fixed temperature of 0.6 and a maximum generation length of 8192 tokens of each round. 
The LLM judge model is implemented with Qwen3-235B-A22B-Instruct-2507, more details are provided in Appendix~\ref{appendix:exp_setup}.

\subsection{Main Result}

\begin{table*}[t]
\centering
\resizebox{0.9999\textwidth}{!}{
\begin{tabular}{lc|cc|cc|cc|cc}
\toprule
\multirow{2}{*}{\textbf{Models}} & \multirow{2}{*}{\textbf{Thinking}} & \multicolumn{2}{c|}{\textbf{Complex Tool Usage}} & \multicolumn{2}{c|}{\textbf{Data Analysis}} & \multicolumn{2}{c|}{\textbf{Deep Search}} & \multicolumn{2}{c}{\textbf{Overall}} \\
\cmidrule(lr){3-4} \cmidrule(lr){5-6} \cmidrule(lr){7-8} \cmidrule(lr){9-10}
 & & Avg@3 & Pass@3 & Avg@3 & Pass@3 & Avg@3 & Pass@3 & Avg@3 & Pass@3 \\
\midrule
\rowcolor{gray!10}\multicolumn{10}{c}{\emph{Open-sourced Models $<$ 100B }} \\
Qwen3-8B & \ding{53} & 2.75 & 7.14 & 0.00 & 0.00 & 6.57 & 12.63 & 4.12 & 8.70 \\
Qwen3-8B & \ding{51} & 4.58 & 9.34 & 1.75 & 5.26 & 7.41 & 13.13 & 5.49 & 10.53 \\
Qwen3-32B & \ding{53} & 6.21 & \underline{13.19} & 1.75 & 5.26 & 8.25 & 16.16 & 6.55 & 13.50 \\
Qwen3-32B & \ding{51} & \underline{6.41} & 12.64 & \underline{2.34} & \underline{7.02} & \underline{9.76} & \underline{19.70} & \underline{7.40} & \underline{15.11} \\
Nex-N1-32B & \ding{53} & \textbf{14.65} & \textbf{26.92} & \textbf{9.36} & \textbf{19.30} & \textbf{18.52} & \textbf{35.35} & \textbf{15.71} & \textbf{29.75} \\
\midrule
\rowcolor{gray!10}\multicolumn{10}{c}{\emph{Open-sourced Models $\ge$ 100B }} \\
Qwen3-235B-A22B-Instruct & \ding{53} & 12.27 & 21.98 & 9.94 & 19.30 & 16.84 & 30.30 & 14.04 & 25.40 \\
Qwen3-235B-A22B-Think & \ding{51} & 7.33 & 12.09 & 2.92 & 7.02 & 13.97 & 22.73 & 9.76 & 16.25 \\
Kimi-K2 & \ding{53} & 19.96 & 34.62 & 16.96 & 31.58 & 22.90 & 39.39 & 20.90 & 36.38 \\
DeepSeek-V3.1 & \ding{53} & \underline{23.26} & \textbf{43.41} & \underline{20.47} & \underline{36.84} & 20.54 & 38.38 & \underline{21.66} & 40.27 \\
Deepssek-V3.2-Exp & \ding{53} & 19.60 & 39.56 & 13.45 & 31.58 & \underline{24.41} & \underline{45.96} & 20.98 & \underline{41.42} \\
GLM-4.6 & \ding{51} & 22.16 & 39.01 & 8.19 & 19.30 & 22.56 & 39.40 & 20.52 & 36.62 \\
Nex-N1-671B & \ding{53} & \textbf{29.30} & \underline{42.86} & \textbf{29.82} & \textbf{49.12} & \textbf{35.53} & \textbf{53.55} & \textbf{32.19} & \textbf{48.52} \\
\midrule
\rowcolor{gray!10}\multicolumn{10}{c}{\emph{Proprietary Models}} \\
Gemini-2.5-Pro & \ding{51} & 21.25 & 36.81 & 12.28 & 24.56 & 22.56 & 36.87 & 20.67 & 35.24 \\
Claude-4.5-Sonnet & \ding{51} & \underline{42.38} & \underline{52.20} & \textbf{39.77} & \underline{50.88} & \underline{31.63} & \underline{42.91} & \underline{37.17} & \underline{47.82} \\
GPT-5 & \ding{51} & \textbf{48.72} & \textbf{67.58} & \underline{39.18} & \textbf{59.65} & \textbf{45.80} & \textbf{65.66} & \textbf{46.15} & \textbf{65.68} \\
\bottomrule
\end{tabular}
}
\caption{
Performance comparison against different LLMs on AgentGym2. For each group, the best result is in \textbf{bold}, and the second-best is \underline{underlined}.
}
\vspace{-5pt}
\label{tab:main_performance}
\end{table*}
\paragraph{AgentGym2 poses significant challenges for current models, especially for open-source series.} 
The evaluation results are illustrated in Table \ref{tab:main_performance}. Both proprietary and open-source models struggle with AgentGym2. Even GPT-5 achieves only 46.15 on Avg@3, while the top-performing open-source agents Nex-N1-671B and DeepSeek-V3.1 achieve merely 32.19 and 21.66, respectively. This demonstrates that AgentGym2 presents a significant challenge for current SOTA models, revealing a substantial gap between the capabilities of existing agents and the demands of real-world deployments.

\paragraph{Model performance varies across different scenarios.}
Most models underperform in Data Analysis scenario compared to other scenarios. For instance, GLM-4.6 achieves 22.16 on Complex Tool Use but only 8.19 on Data Analysis, a drop of 13.97 points. 
Claude-4.5-Sonnet and the Nex-N1 series maintain much more stable performance in Data Analysis, which can be attributed to their strong agentic coding capabilities, empowering them to parsie complex dependencies between data entries and multimodal files, thereby leading to higher task success rates.

\paragraph{Agentic post-training improves model performance.}
Nex-N1-32B (post-trained from Qwen3-32B) and Nex-N1-671B (post-trained from DeepSeek-V3.1) outperform their respective base models by approximately 9.16 and 10.5 in Avg@3. These gains underscore the efficacy of agentic post-training and suggest that it significantly narrows the performance gap with proprietary systems.

\subsection{Scaling Trends with Model size, Interaction Rounds, and Budget}

In Figure \ref{fig:params_performance}, we explore the effect of model performance across three key dimensions: parameter scale, total cost, and the average number of interaction rounds.

\paragraph{Performance exhibits a clear upward trend correlated with model scale.}
Within the Qwen3 series, the 8B, 32B, and 235B (non-thinking) variants achieve scores of 8.70, 10.53, and 25.40, respectively. This suggests that scaling model size remains an effective dimension for improving performance and is worth further investigation.

\paragraph{High-performing models tend to exhibit a moderate number of interaction rounds per task.} 
Top-tier models, such as GPT-5, Claude-4.5-Sonnet, and Nex-N1-671B, typically operate within 20–30 interaction rounds per task. Notably, reasoning-oriented (thinking-mode) variants require fewer interactions than non-thinking modes, as they prioritize internalized deliberation over frequent external tool invocation. Specifically, Qwen3-235B-A22B-Thinking-2507 exhibits the lowest interaction frequency among all competitors; however, its performance lags behind the instruct version by 5.31 points. This performance gap indicates that balancing reasoning and interaction is crucial for optimizing performance in real-world tasks.

\paragraph{Cost-performance Trade-offs.} 

Top-performing models typically come with high costs, and practitioners must balance performance, efficiency, and budget. This trade-off suggests that current agents still fall short of real-world production needs, where both strong performance and cost efficiency are essential.

\section{Analysis and Discussion}
\label{sec:analysis}
\begin{wrapfigure}{r}{8cm}
\vspace{-40pt}
    \includegraphics[width=0.99\linewidth]{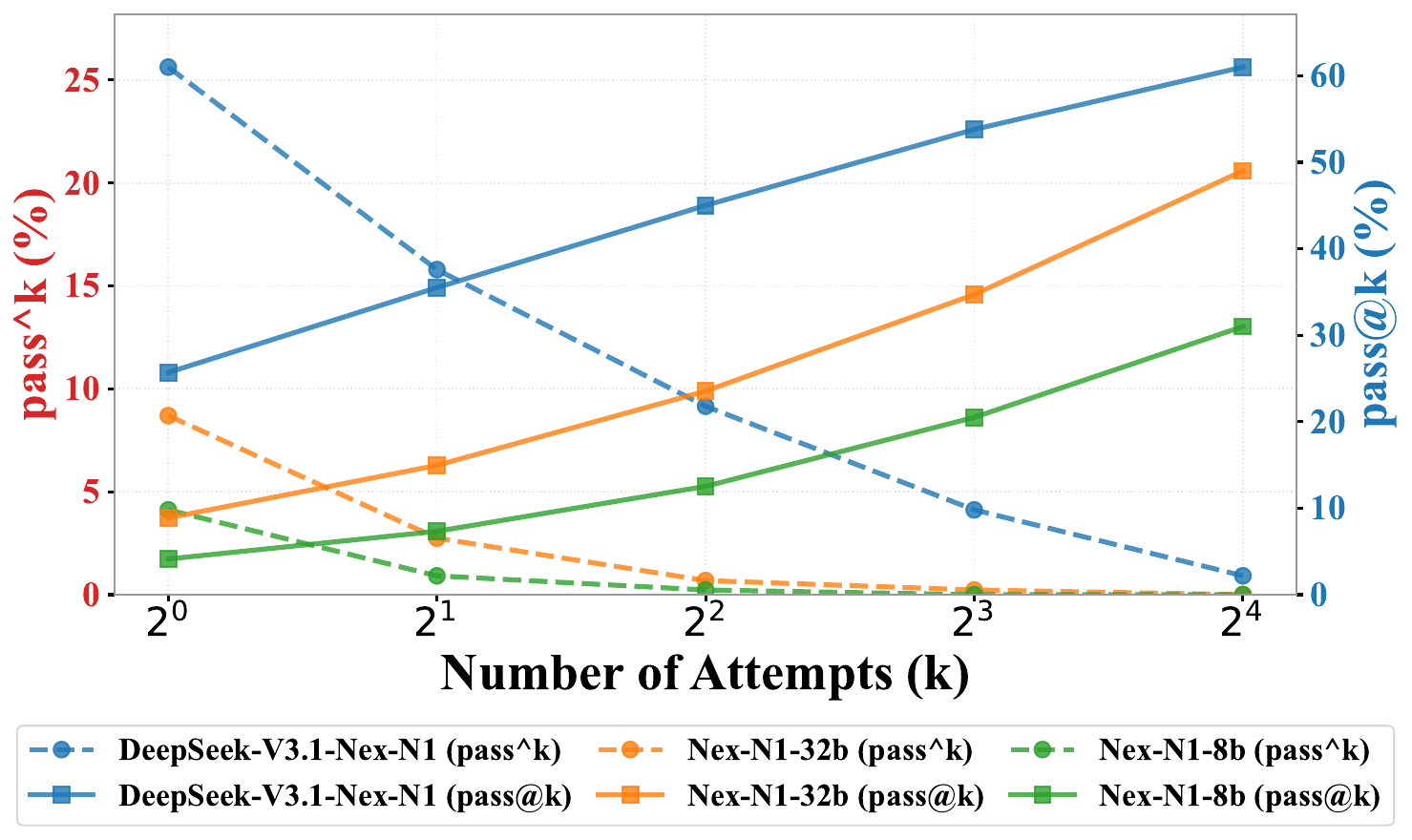 }
    \centering
 	\caption{Pass@k and Pass\textasciicircum k performance on AgentGym2. Pass@k measures the probability that at least one correct solution appears among the sampled trajectories, whereas Pass\textasciicircum k measures the consistency that all of the k sampled trajectories are correct.}
  \label{fig:test_scaling}
  \vspace{-20pt}
\end{wrapfigure} 

\paragraph{LLM‑as‑judge reliability.}

We measure human–LLM agreement during our evaluation. Specifically, we manually inspected all outputs from four representative models—Gemini-2.5-Pro, GPT-5, Claude-4.5-Sonnet, and Nex-N1-671B—and compared each agent output against the reference answer. The overall agreement between the LLM judge and human verification is 98.9\%, supporting the reliability of the LLM-as-judge.

\begin{wrapfigure}{r}{8cm}
\vspace{-40pt}
    \includegraphics[width=0.99\linewidth]{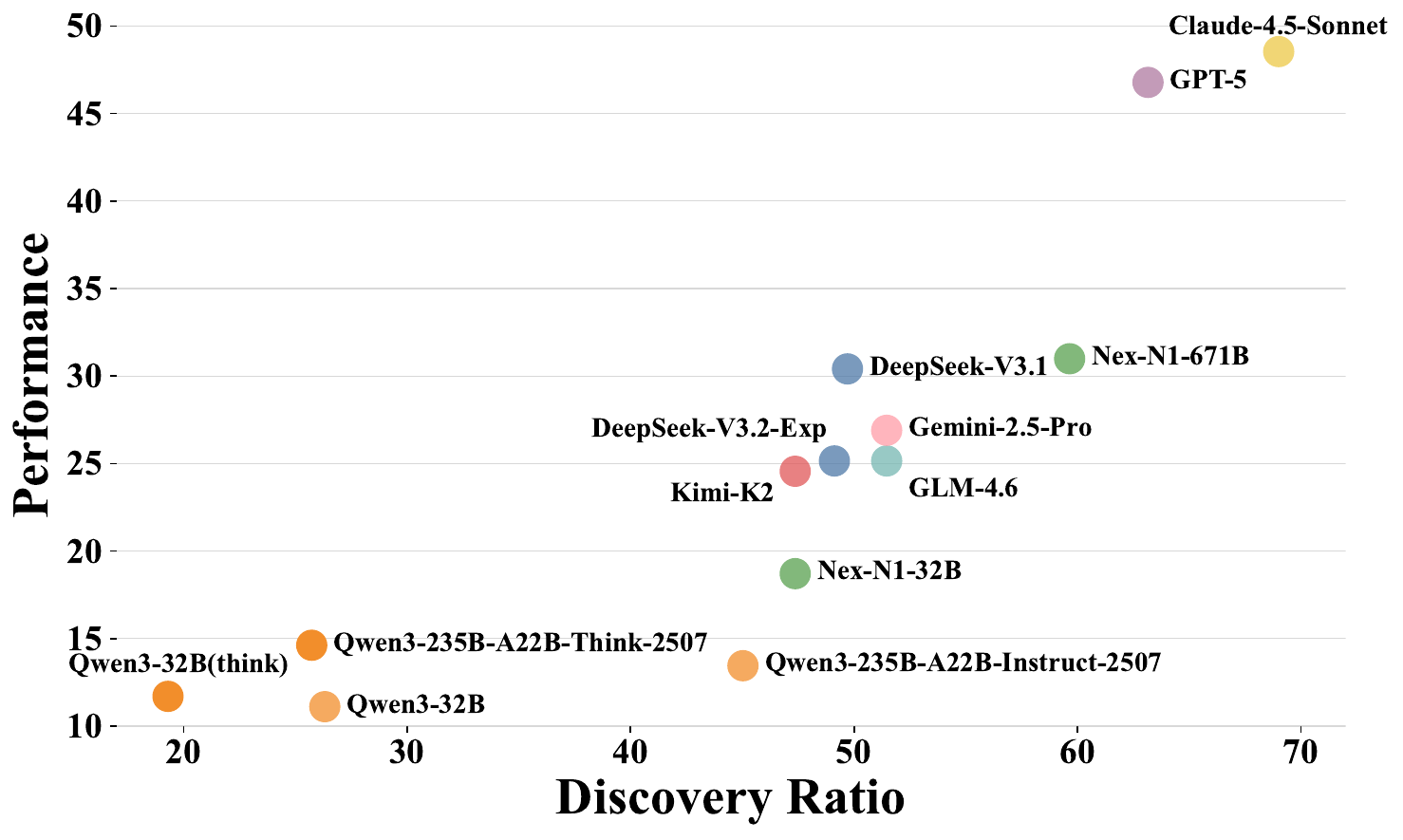}
    \centering
 	\caption{Relationship between discovery ratio and performance on complex tool using. The discovery ratio denotes the proportion of tasks in which agents successfully discover relevant online tools.}
  \vspace{2mm}
  \label{fig:tool_discovery_ratio}
   \vspace{-20pt}
\end{wrapfigure}

\paragraph{Pass@k and Pass\textasciicircum k performance.}

We investigate the effect of test-time compute scaling on a subset of AgentGym2 (100 instances) using the Nex-N1 series models. Specifically, we sample multiple trajectories for each query and report their performance under sampling budget $N$.

As shown in Figure \ref{fig:test_scaling}, increasing test-time compute improves Pass@k, indicating that more tasks can be solved. For example, the high-capacity Nex-N1-671B gains up to \textasciitilde 30\% with larger sampling budgets. In contrast, the Pass\textasciicircum k metric declines sharply because it requires success in all $N$ samples, highlighting the model’s limited stability and robustness in deployment.

\paragraph{The importance of tool discovery and downstream procedures.}

In AgentGym2, tool discovery is a key evaluation focus, and we analyze this subset in depth. We define the discovery ratio as the proportion of agents that successfully identify relevant specialized online tools. As shown in Figure \ref{fig:tool_discovery_ratio}, discovery ratio correlates strongly with task performance, indicating that tool discovery is a prerequisite for success. Yet performance remains limited even when the right tools are found, suggesting that downstream abilities—such as tool composition and correct invocation—are equally critical and underscoring the need for end-to-end competence.

\paragraph{Data wrangling poses significant challenges.}

In AgentGym2, we argue that effective data analysis requires both data wrangling and downstream analysis. Wrangling tests robustness through dataset exploration, validation, and automatic repair (e.g., missing values and schema inconsistencies), while the analysis stage requires careful reasoning and insight extraction to answer the task.

\begin{wrapfigure}{r}{8.5cm}
\vspace{-10pt}
    \includegraphics[width=0.99\linewidth]{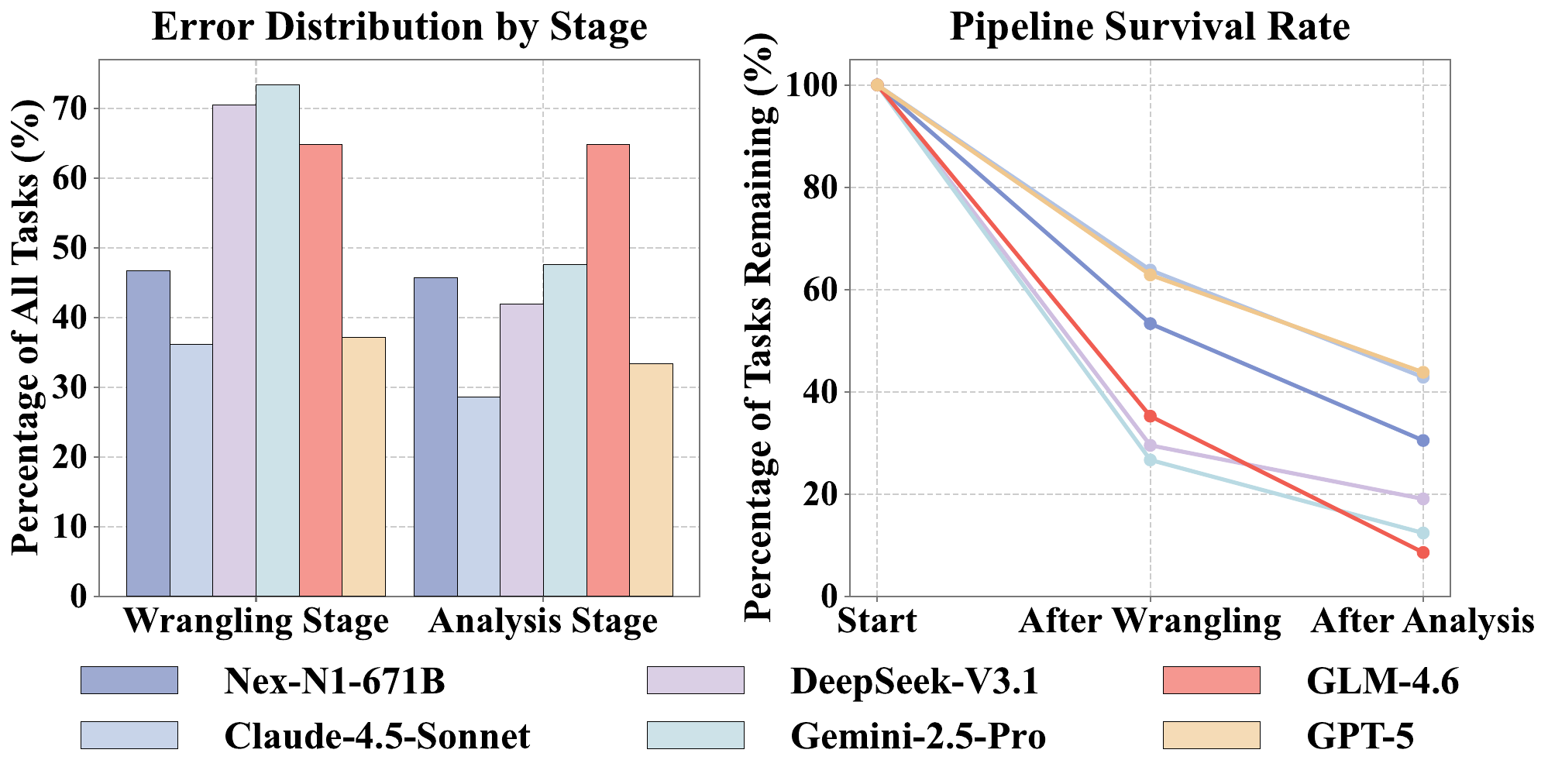}
    \centering
 	\caption{
Error distribution (left) and survival rates (right) across the data wrangling and analysis stages.
  }
  \label{fig:da_pipeline}
\end{wrapfigure}

Hence, we examine the error distribution and survival rates across the data wrangling and analysis stages, as illustrated in Figure~\ref{fig:da_pipeline}. The left panel reveals that data wrangling constitutes the primary bottleneck in data analysis scenarios. 
The right panel shows that the survival rate drops most sharply after data wrangling, suggesting that future agent development should account for complex real-world environments and full end-to-end workflows.

\paragraph{Impact of noisy information on deep search performance.}

\begin{wrapfigure}{r}{8.5cm}
\vspace{-20pt}
    \includegraphics[width=0.99\linewidth]{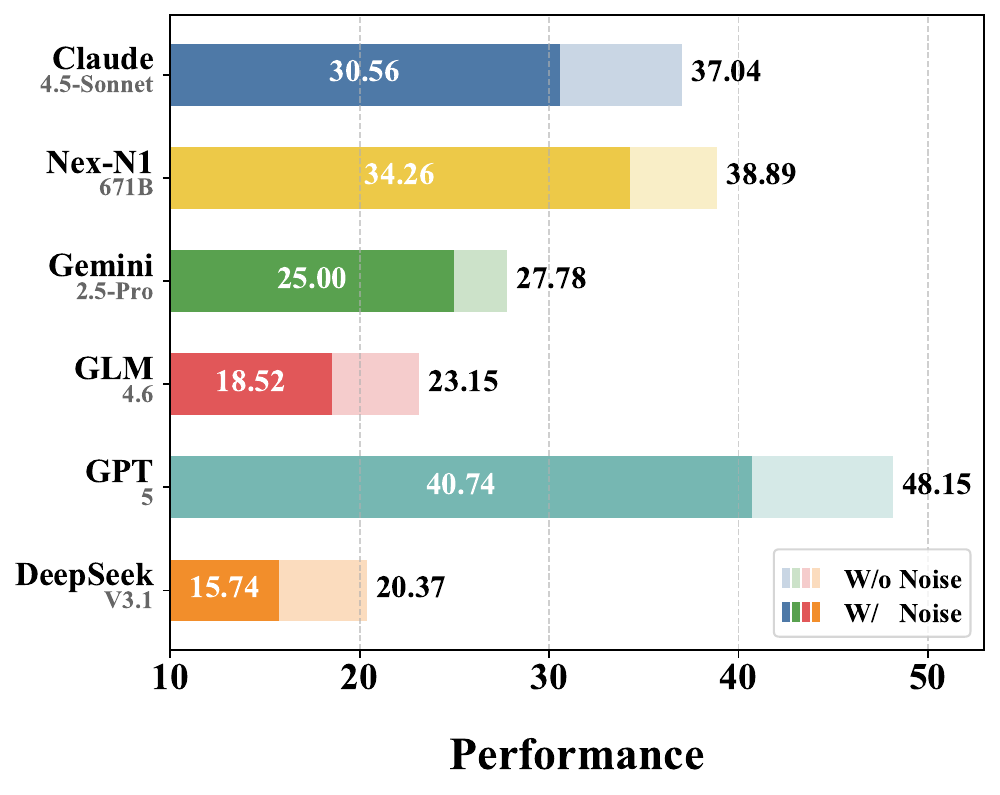}
    \centering
 	\caption{
Effect of the added noise in query on deep search.
  }
  \label{fig:search_analysis}
  \vspace{-25pt}
\end{wrapfigure}

Real users’ deep search queries are often noisy and may even reflect memory biases \citep{DBLP:journals/corr/abs-2409-00557,DBLP:journals/corr/abs-2508-04183,li2024dmqrrag}, and agent performance in such settings is crucial for user experience. To simulate this, we inject noise into otherwise correct queries and evaluate models with and without the added noise. As shown in Figure \ref{fig:search_analysis}, introducing noisy information causes substantial performance drops across all models, with Claude-4.5-Sonnet decreasing by 6.4\% and GPT-5 by 7.4\%. This indicates that current agents remain insufficiently robust to noisy information in real-world queries.

\paragraph{Failure mode analysis.}
We conduct a fine-grained error analysis across all scenarios, using GPT-5 to categorize failures according to the taxonomy in Table \ref{tab:tool_definitions_detailed} in Appendix \ref{appendix:failure_mode}, where we provide detailed descriptions of each error type.

\begin{figure*}[t]
    \includegraphics[width=0.99\linewidth]{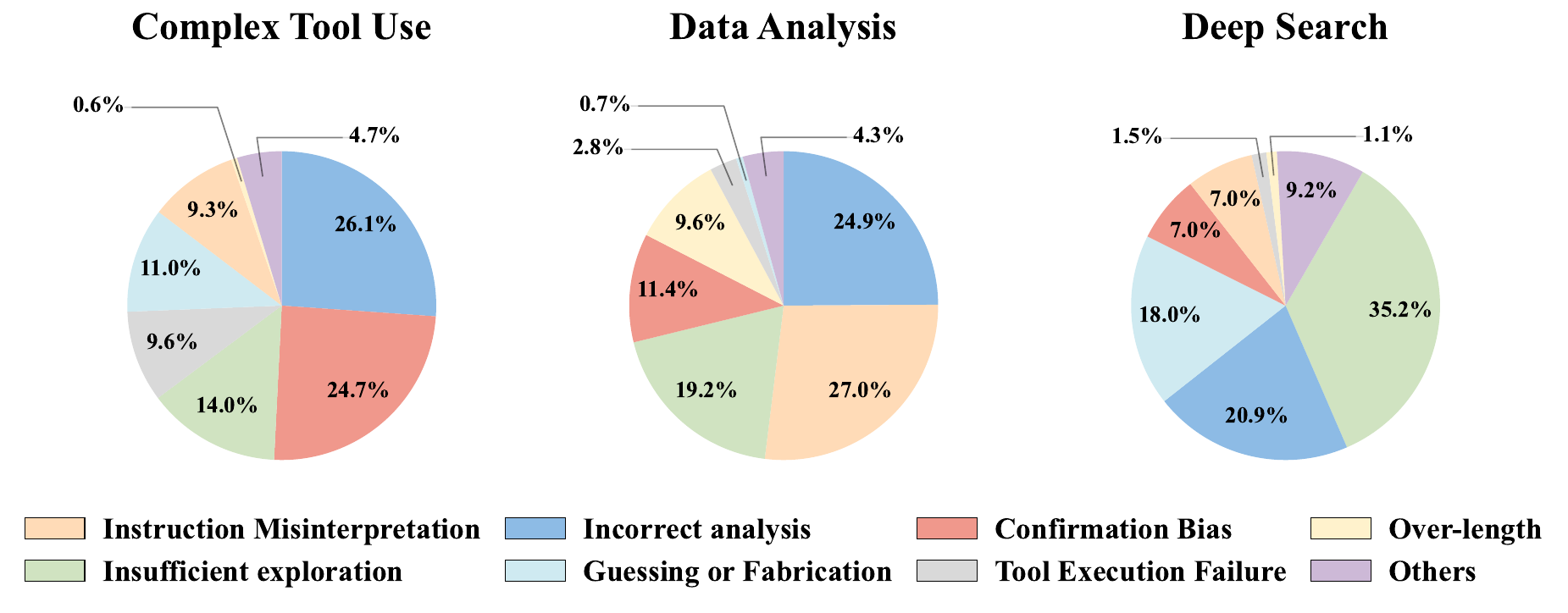}
    \centering
 	\caption{
Failure model analysis across all scenarios, where incorrect analysis and insufficient exploration consists of the main error modes.
  }
  \vspace{-2mm}
  \label{fig:failure_mode_analysis}
\end{figure*}

As shown in Figure \ref{fig:failure_mode_analysis}, overall, the model exhibits different failure patterns across tasks; however, incorrect analysis (24.0\% on average) and insufficient exploration (22.8\% on average) consistently emerge as the leading error modes across scenarios. This suggests that under real-world conditions, the model’s reasoning/analysis capability and its ability to explore the open world remain key weaknesses.
In complex tool use, failures are primarily characterized by confirmation bias (24.7\%), where the model is overconfident and relies on internal priors rather than actively discovering or invoking external tools for verification. In data analysis, instruction misinterpretation is the dominant failure mode (27.0\%), indicating difficulty in maintaining precise alignment with fine-grained task constraints. In deep search, errors are largely driven by insufficient exploration (35.2\%), with the model prematurely terminating search trajectories in noisy and ambiguous environments. Collectively, these patterns reveal limited robustness in sustaining multi-turn decision processes and insufficient utilization of environmental feedback, which remain key bottlenecks for deploying LLM agents in complex, real-world tasks.

\begin{figure*}[t]
\includegraphics[width=0.8\textwidth]{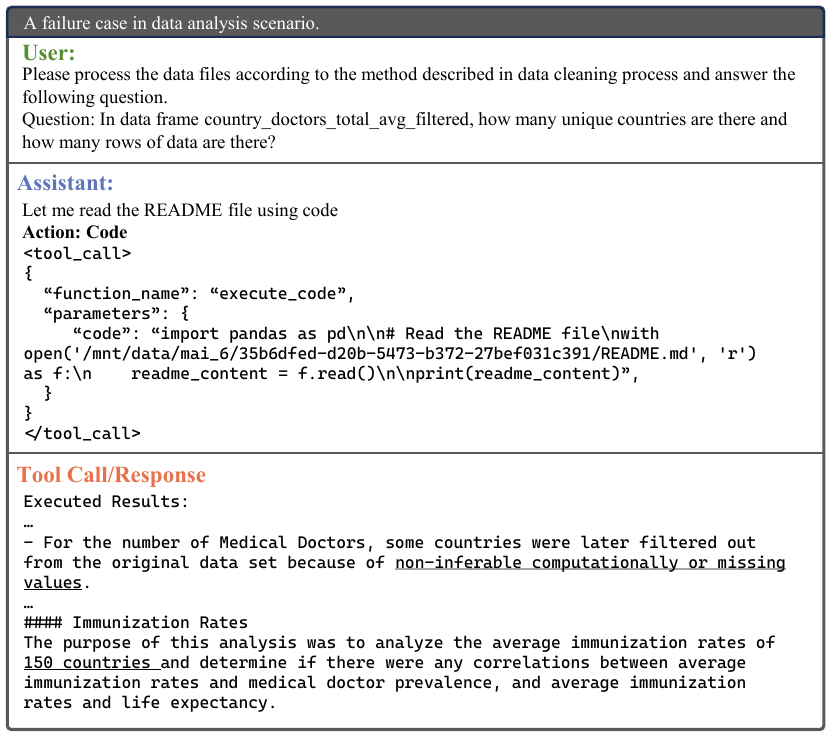}
\centering
\caption{
Part 1/2 of the data analysis case study: The task requires the model to perform the corresponding cleaning and analysis based on the problem background and cleaning rules provided in the README.
}
\vspace{-1mm}
\label{fig:case_study31}
\end{figure*}

We have presented a failure case of data analysis, where model fail to sufficiently explore the whole dataset, as illustrated in Figure \ref{fig:case_study31} and Figure \ref{fig:case_study32}. We further present detailed case studies of other representative failure modes in Appendix \ref{appendix:failure_cases} (Figures \ref{fig:case_study11}–\ref{fig:case_study23}).

\begin{figure*}[t]
\includegraphics[width=0.8\textwidth]{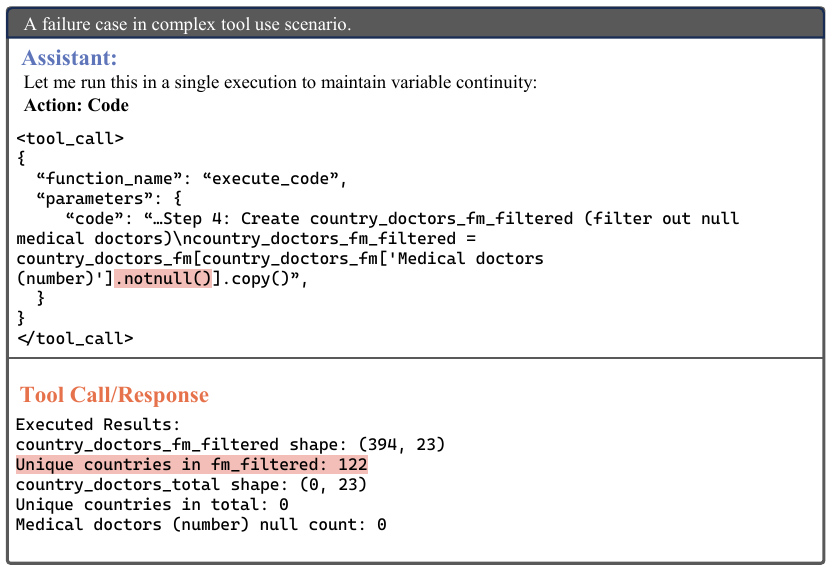}
\centering
\caption{
Part 2/2 of the data analysis case study: However, due to the model’s \textbf{Insufficient Exploration}, it has limited understanding of the dataset itself, leading to a misinterpretation of the task during cleaning and the direct removal of all null data. This \textbf{Instruction Misinterpretation} directly results in the final incorrect answer.
}
\vspace{-1mm}
\label{fig:case_study32}
\end{figure*} 

\paragraph{Ablation study on de-idealized factors.} 
We select subsets and conduct controlled ablations on key de-idealized factors to quantify their impact on model performance.
\begin{itemize}
    \item \textbf{Hidden-information retrieval}: In some tasks, we place important information or clues inside files, and the agent must read the files and filter out irrelevant content (distractors). Here, we ablate this factor by adopting an idealized setting in which we directly provide the agent with this information.
\item \textbf{Tool discovery}: In some tasks, agents need to discover new tools to solve the task. Here, we ablate this factor by adopting an idealized setting where the required tools are provided to the agent directly.
\item \textbf{Ambiguity/noise with bias}: In some tasks, we include noisy content, such as bias introduced by a user’s memory. Here, we ablate this factor by adopting an idealized setting that removes such noise and biases.
\end{itemize}
The results in Table~\ref{tab:ablation_performance} show that performance improves substantially after ablation, indicating that these common real-world non-ideal factors are key contributors to performance degradation. By explicitly incorporating them, AgentGym2 better reflects the requirements of real deployment. 

\begin{table*}[t]
\centering
\resizebox{0.98\textwidth}{!}{%
\begin{tabular}{lcccccc}
\toprule
\multirow{2}{*}{\textbf{Models}} & \multicolumn{2}{c}{\textbf{Hidden-information retrieval}} & \multicolumn{2}{c}{\textbf{Tool Discovery}} & \multicolumn{2}{c}{\textbf{Ambiguity/noise with bias}} \\
\cmidrule(lr){2-3} \cmidrule(lr){4-5} \cmidrule(lr){6-7}
& idealized & de-idealized & idealized & de-idealized & idealized & de-idealized \\ \midrule
Gemini-2.5-Pro & 33.33 & 18.67 & 33.33 & 26.90 & 27.78 & 25.00 \\
GPT-5          & 61.67 & 49.60 & 53.57 & 46.78 & 48.15 & 40.74 \\ \bottomrule
\end{tabular}
}
\caption{Model performance in de-idealized factor settings. All results are reported in Avg@3.}
\label{tab:ablation_performance}
\end{table*}

\section{Conclusion}

In this work, we introduced AgentGym2, a new framework to evaluate language agents in de-idealized real-world settings. 
Beyond reasoning and planning, AgentGym2 measures agents’ ability to execute end-to-end procedures, discover tools via exploration, compose tools for unseen tasks, and remain robust to noisy and underspecified information.
We implement the whole framework by equipping models with basic and composable tools, enabling isolated, parallel, and scalable interactions. 
Our extensive experiments across 27 diverse domains show that AgentGym2 remains highly challenging for current state-of-the-art models. We hope this work will help the community build more robust agents that can meaningfully improve real-world productivity.



\newpage
\bibliography{main}

\appendix

\section{Supplement to Benchmark}

\subsection{Data Construction}

\paragraph{Complex tool using scenario.}

In this scenario, we have a subset of questions that originate from volunteers. We have collected a set of real questions and discussions from 10 volunteers. The volunteers are primarily undergraduate students and above, spanning multiple academic disciplines. During the data collection process, we first introduce the definition of tool discovery to the volunteers—for example, using online services such as WolframAlpha for mathematical formula computation, or accessing official websites to download datasets. We then ask the volunteers to provide recent examples in which they have used tool discovery, along with the corresponding usage tasks. The full instruction is "Please refer to the above definition of tool discovery and, based on your own experience and recent work, provide some scenarios in which tool discovery is used along with the corresponding objectives. Note that these data will be used for benchmark construction, so please do not include any private or sensitive information". From these submissions, we select publicly available data covering a wide range of domains, including academia, daily life, and sports, and conducted subsequent adaptation. These data ultimately form 57 instances for our tool discovery task.

We also involve authors who have a strong understanding of recent deep search work to enhance the existing datasets. The authors are instructed as follows: “Please refer to recent deep search work on the decomposition of entities and relations to break down the existing datasets, and search for or create relevant attachment files to replace key entities, ultimately forming new problems.”

\paragraph{Data analysis scenario.}

In this scenario, 4 authors extensively explore and collect data from GitHub and Kaggle, with a particular focus on datasets related to complex data cleaning. Our authors are instructed as "please use keywords such as data cleaning and data wrangling to search for relevant datasets on websites like GitHub and Kaggle. Priority should be given to complex datasets that come with existing reproduction files and execution traces". We prioritize datasets that provided reproducibility files and conduct manual data inspection and reproduction workflows to ensure the reliability of the entire process. Based on these procedures, we collaborate with 4 volunteers with a background in statistics to create a data-cleaning README for each dataset, requiring "complete, unique, and well-defined cleaning procedures that require sufficient exploration of the dataset itself to be fully understood". We ensure that the cleaning pipeline was unique, while still requiring exploratory data analysis (EDA) of the data itself to complete the process. These data finally form the final 57 instances.

\paragraph{Deep search scenario.}

In this scenario, we collect real question data from social-media users (e.g. Reddit, Xiaohongshu). We prioritize queries that contained hedging adverbs and where the asker had already accepted a correct answer. We manually verify the responses from other users to ensure that only the accepted answer significantly matched the content of the question. The collected data are then standardized—for example, converting “10 years ago” to “before 2015” and removing certain meme emojis. These data ultimately form 48 instances for our Ambiguity-with-bias task.

\subsection{Benchmark statistics}

Table \ref{tab:attachment_detail} details the number and type of attachment files in our benchmark.

\begin{table}[htbp]
\centering
\resizebox{0.6\linewidth}{!}{
\begin{tabular}{lccc}
\toprule
\textbf{Scenario} & \textbf{File Type} & \textbf{Nums} & \textbf{Total} \\
\midrule

\multirow{9}{*}{Complex Tool Usage}
 & csv  & 204 & \multirow{9}{*}{409}  \\
 & json & 169 &  \\
 & jpg  & 15  &  \\
 & png  & 9   &  \\
 & mp4  & 5   &  \\
 & txt  & 3   &  \\
 & pdf  & 2   &  \\
 & svg  & 1   &  \\
 & py   & 1   &  \\
\midrule
\multirow{11}{*}{Data Analysis}
 & csv  & 181 &  \multirow{11}{*}{243} \\
 & md   & 26  &  \\
 & json & 20  &  \\
 & txt  & 6   &  \\
 & py   & 3   &  \\
 & yml  & 2   &  \\
 & pdf  & 1   &  \\
 & tsv  & 1   &  \\
 & html & 1   &  \\
 & yaml & 1   &  \\
 & log  & 1   &  \\
\midrule
Deep Search & --- & --- & --- \\
\midrule
Total & --- & --- & 652 \\
\bottomrule
\end{tabular}
}
\caption{Distribution of attachment file types across different scenarios on AgentGym2.}
\label{tab:attachment_detail}
\end{table}

\section{Details on toolbox}

The detailed definition of 27 tools across web browsing, information retrieval, file processing, multimodal understanding and code execution are shown in Table \ref{tab:tool_definitions_detailed}. In practice, all tools are encapsulated as functions, and their definitions along with the input–output formats of parameters are passed to the agent via the OpenAI tool schema.

\begin{table*}[t]
\centering
\small
\setlength{\tabcolsep}{4pt} 
\renewcommand{\arraystretch}{1.5} 
\setlength{\parskip}{0pt}
\setlength{\parindent}{0pt}
\resizebox{0.7\textwidth}{!}{
\begin{tabular}{
  l |
  >{\raggedright\arraybackslash}p{6cm} |
  >{\raggedright\arraybackslash}p{3cm} |
  >{\raggedright\arraybackslash}p{2.5cm}
}
\toprule
\textbf{Name} & \textbf{Description} & \textbf{Input} & \textbf{Output} \\
\midrule

\multicolumn{4}{l}{\cellcolor{gray!15}\textbf{\textit{- Web Browsing}}} \\
scrape & Fetch and extract the main textual content from a given webpage. & [text] URL. [text] Query & [text] Extracted webpage content \\
browse\_url & A powerful toolkit which can simulate the browser interaction to solve the task which needs multi-step actions. & [text] Task prompt. [text] Start URL. [int] Round limit & [text] Task completion result \\
\midrule

\multicolumn{4}{l}{\cellcolor{gray!15}\textbf{\textit{- Supported Browser Actions (used within \texttt{browse\_url})}}} \\
fill\_input\_id & Fill an input field and submit the text. & [text/int] Element ID. [text] Input text & Page updated \\
click\_id & Click an element on the page. & [text/int] Element ID & Page updated \\
hover\_id & Hover over an element to trigger UI changes. & [text/int] Element ID & Page updated \\
download\_file\_id & Download a file from the page. & [text/int] Element ID & [file] Local file path \\
scroll\_to\_top/bottom & Scroll to the top/bottom of the page. & --- & Page updated \\
scroll\_up/down & Scroll upward/downward within the page. & --- & Page updated \\
back & Navigate back to the previous page. & --- & Page updated \\
stop & Stop the browsing process and finalize the result. & --- & [text] Final result \\
get\_url & Retrieve the current page URL. & --- & [text] Current URL \\
find\_text\_on\_page & Find and scroll to specific text on the page. & [text] Search text & Page scrolled \\
visit\_page & Navigate directly to a specified URL. & [text] URL & Page updated \\
click\_blank\_area & Click a blank area to remove focus from active elements. & --- & Page updated \\
ask\_question\_about\_video & Ask a question about video content on the page. & [text] Question & [text] Answer \\
\midrule

\multicolumn{4}{l}{\cellcolor{gray!15}\textbf{\textit{- Information Retrieval}}} \\
search\_wiki & Search Wikipedia for a specific entity and return its factual summary. & [text] Entity name & [text] Entity summary \\
search\_serper & Perform Google web or news search and return top search results. & [text] Query and parameters & [text] Search results \\
google\_scholar & Retrieve relevant academic publications from Google Scholar. & [text] Academic query & [text] Scholarly results \\
\midrule

\multicolumn{4}{l}{\cellcolor{gray!15}\textbf{\textit{- File Processing}}} \\
extract\_document\_content & Extract the content of a document and return the processed text. & [file] Document path & [text] Extracted text \\
extract\_excel\_content & Extract detailed cell and sheet information from an Excel file. & [file] Excel path & [text] Structured content \\
\midrule

\multicolumn{4}{l}{\cellcolor{gray!15}\textbf{\textit{- Multimodal Understanding}}} \\
audio2text & Transcribe audio content into text. & [audio] File or URL & [text] Transcription \\
ask\_question\_about\_audio & Answer questions based on the semantic understanding of audio. & [audio] File. [text] Question & [text] Answer \\
image\_to\_text & Generate a textual description of an image. & [image] File. [text] Prompt & [text] Description \\
ask\_question\_about\_image & Answer questions about image content with optional instructions. & [image] Image file or URL. [text] Question. [text] System prompt & [text] Answer \\

\multicolumn{4}{l}{\cellcolor{gray!15}\textbf{\textit{- Code Execution}}} \\
execute\_code & Execute a code snippet in a stateless environment and return the result. & [text] Source code & [text] Execution result \\

\bottomrule
\end{tabular}
}
\caption{Detailed definition and prompts of tools across categories on AgentGym2.}
\label{tab:tool_definitions_detailed}
\end{table*}

\section{Experimental Setup Details}
\label{appendix:exp_setup}
\subsection{LLM Implementation Details}

For open-source models under 100B parameters, we deploy them locally on NVIDIA A100 GPUs and provide an OpenAI-compatible server using the vllm library\cite{kwon2023efficient}.
For other models, we access the official providers and conduct standardized testing through their official APIs.

\subsection{Other Implementation Details}

To prevent potential data contamination, we manually disable the models’ access to the relevant websites when testing queries sourced from real online data.

\section{Prompt Details}
\label{sec:prompt_details}

\subsection{System Prompt}

The system prompt details, adapted from owl \cite{hu2025owl}, for our evaluation is shown in Figure \ref{fig:system_prompt}. To make a fair comparison between different models, we use this same prompt to every query and to every model. This system prompt, together with tool definitions, are prompted to the agent first before the task query.

\begin{figure*}[t]
\centering
\begin{tcolorbox}[
  colback=white,
  colframe=black,
  arc=3mm,
  boxrule=0.8pt,
  title=\centering System Prompt,
  fonttitle=\bfseries,
]
===== RULES OF ASSISTANT =====

You are a helpful assistant. You have to utilize your available tools to solve the task I assigned.

Please note that our overall task may be very complicated. Here are some tips that may help you solve the task:

\textless tips\textgreater

- If one way fails to provide an answer, try other ways or methods. The answer does exists.

- If the search snippet is unhelpful but the URL comes from an authoritative source, try visit the website for more details.

- When looking for specific numerical values (e.g., dollar amounts), prioritize reliable sources and avoid relying only on search snippets.

- When solving tasks that require web searches, check Wikipedia first before exploring other websites.

- When trying to solve math problems, you can try to write python code and use sympy library to solve the problem.

- Always verify the accuracy of your final answers!

- Do not be overly confident in your own knowledge.

- You can use browser tools to access full browser functionality, including rich interactions, file downloads, and page visits, etc.

- After writing codes, do not forget to run the code and get the result.

- When a tool fails to run, never assume that it returns the correct result.

- Do not attempt to read or parse any files directly.

- Always read all README files before taking action.

\textless/tips\textgreater

Here are some hint about the final answer after solving the whole task:

\textless hint\textgreater

Your final answer must be enclosed by \textless final\_answer\textgreater \textless/final\_answer\textgreater.

Your final answer must be output exactly in the format specified by the question.

\textless/hint\textgreater

Here is the overall task:

\{task\}

Please solve this task step by step. Never forget the task!
\end{tcolorbox}
\caption{System Prompt of AgentGym2.}
\label{fig:system_prompt}
\end{figure*}

\begin{figure*}[t]
\centering
\begin{tcolorbox}[
  colback=white,
  colframe=black,
  arc=3mm,
  boxrule=0.8pt,
  title=\centering LLM-as-Judge Prompt for Complex Tool Usage,
  fonttitle=\bfseries,
]
Based on the given question, standard answer, and model-predicted answer, evaluate whether the model's response is correct. Your task is to classify the result as: [CORRECT] or [INCORRECT].

First, we'll list examples for each category, then you'll evaluate a new question's predicted answer.
Here are examples of [CORRECT] responses:

Question: What are the names of Barack Obama's children?

Standard Answer: Malia Obama and Sasha Obama

Model Prediction 1: Malia Obama and Sasha Obama

Model Prediction 2: Malia and Sasha

Model Prediction 3: Most would say Malia and Sasha, but I'm not sure, I should verify

Model Prediction 4: Barack Obama has two daughters, Malia Ann and Natasha Marian, commonly known as Malia Obama and Sasha Obama.

These responses are all [CORRECT] because they:

    - Fully include the important information from the standard answer.

    - Don't contain any information that contradicts the standard answer.
    
    - Focus only on semantic content; language, capitalization, punctuation, grammar, and order aren't important.
    
    - Vague statements or guesses are acceptable as long as they include the standard answer and don't contain incorrect information or contradictions.
    
Here are examples of [INCORRECT] responses:

Question: What are the names of Barack Obama's children?

Standard Answer: Malia Obama and Sasha Obama

Model Prediction 1: Malia

Model Prediction 2: Malia, Sasha and Susan or Sasha Obama or Malia Obama, or Natasha Marian, or Einstein

Model Prediction 3: While I don't know their exact names, I can tell you Barack Obama has two children.

Model Prediction 4: You might be thinking of Betsy and Olivia. But you should verify the details with the latest references. Is that the correct answer?

Model Prediction 5: Barack Obama's children

These responses are all [INCORRECT] because they:

    - Contain factual statements that contradict the standard answer.
    
    - Are empty or merely repeat the question.
    
    - Enumerate multiple answers or repeat the answer.

\vspace{\baselineskip}
Below is a new question example. Please reply with only [CORRECT] or [INCORRECT], without apologies or corrections to your own errors, just evaluate the answer.

Question:  \{question\}

Standard Answer: \{correct\_answer\}

Predicted Answer: \{response\}

Evaluate this new question's predicted answer as one of the following:

A. [CORRECT]

B. [INCORRECT]

Return only the option representing [CORRECT] or [INCORRECT], i.e., just return A or B, without adding any other text.

\end{tcolorbox}
\caption{LLM-as-Judge prompt on Complex Tool Usage task.}
\label{fig:complextoolusage_prompt}
\end{figure*}

\begin{figure*}[t]
\centering
\begin{tcolorbox}[
  colback=white,
  colframe=black,
  arc=3mm,
  boxrule=0.8pt,
  title=\centering LLM-as-Judge Prompt for Data Analysis,
  fonttitle=\bfseries,
]
You are an evaluation assistant. Please determine if the predicted answer is equivalent to the labeled answer.

Question: \{question\}

Labeled Answer: \{correct\_answer\}

Predicted Answer: \{response\}

Did the model give an answer **equivalent** to the labeled answer? Please respond with "Correct" if they are equivalent, or "Incorrect" if they are not equivalent. Do not include any other text.
\end{tcolorbox}
\caption{LLM-as-Judge prompt on Data Analysis task.}
\label{fig:dataanalysis_prompt}
\end{figure*}

\begin{figure*}[t]
\centering
\resizebox{0.92\textwidth}{!}{
\begin{tcolorbox}[
  colback=white,
  colframe=black,
  arc=3mm,
  boxrule=0.8pt,
  title=\centering LLM-as-Judge Prompt for Deep Search,
  fonttitle=\bfseries,
]
Your job is to look at a question, a gold target, and a predicted answer, and then assign a grade of either ["CORRECT", "INCORRECT", "NOT\_ATTEMPTED"].\par
First, I will give examples of each grade, and then you will grade a new example.\par
\par
The following are examples of CORRECT predicted answers.\par

Question: What are the names of Barack Obama's children?\par
Gold target: Malia Obama and Sasha Obama\par
Predicted answer 1: sasha and malia obama\par
Predicted answer 2: most people would say Malia and Sasha, but I'm not sure and would have to double check.\par

These predicted answers are all CORRECT because:\par
    - They fully contain the important information in the gold target.\par
    - They do not contain any information that contradicts the gold target.\par
    - Only semantic meaning matters; capitalization, punctuation, grammar, and order don't matter.\par
    - Hedging and guessing are permissible, provided that the gold target is fully included and the response contains no incorrect information or contradictions.\par
\par
The following are examples of INCORRECT predicted answers.\par

Question: What are the names of Barack Obama's children?\par
Gold target: Malia and Sasha\par
Predicted answer 1: Malia.\par
Predicted answer 2: Malia, Sasha, and Susan.\par
Predicted answer 3: Barack Obama does not have any children.\par
Predicted answer 4: While I don't know their exact names, I can tell you that Barack Obama has three children.\par

These predicted answers are all INCORRECT because:\par
    - A factual statement in the answer contradicts the gold target. Incorrect statements that have some hedging (e.g., "it is possible that", "although i'm not sure, i think") are also considered incorrect.\par
\par
The following are examples of NOT\_ATTEMPTED predicted answers.\par

Question: What are the names of Barack Obama's children?\par
Gold target: Malia and Sasha\par
Predicted answer 1: I don't know.\par
Predicted answer 2: I need more context about which Obama you are talking about.\par
Predicted answer 3: Without researching the web, I cannot answer this question. However, I can tell you that Barack Obama has two children.\par

These predicted answers are all NOT\_ATTEMPTED because:\par
    - The important information in the gold target is not included in the answer.\par
    - No statements in the answer contradict the gold target.\par
\par
Here is a new example. Simply reply with either CORRECT, INCORRECT, NOT ATTEMPTED. Don't apologize or correct yourself if there was a mistake; we are just trying to grade the answer.\par

Question: \{question\}\par
Gold target: \{correct\_answer\}\par
Predicted answer: \{response\}\par

\par
Grade the predicted answer of this new question as one of:\par
A: CORRECT\par
B: INCORRECT\par
C: NOT\_ATTEMPTED\par
\par
Just return the letters "A", "B", or "C", with no text around it.
\end{tcolorbox}
}
\caption{LLM-as-Judge prompt on Deep Search task.}
\label{fig:deepsearch_prompt}

\end{figure*}

\subsection{LLM-as-Judge Prompt}

The LLM-as-Judge prompt specifications are illustrated in Figure \ref{fig:complextoolusage_prompt}, Figure \ref{fig:dataanalysis_prompt}, and Figure \ref{fig:deepsearch_prompt}. To ensure reliable evaluation performance, we adopt three scenario-specific prompts, adapted from Tongyi-DeepResearch\cite{DBLP:journals/corr/abs-2510-24701}.

\section{Failure Case Study}

\subsection{Failure mode analysis}
\label{appendix:failure_mode}
To further investigate the models’ failure modes, we manually examine and summarize several types of errors. We use GPT-5 to annotate the error types for incorrect data. We provide a possible correct reference answer trajectory, along with our definitions for each error type, as shown in Table \ref{tab:failure_type}.

\begin{table*}[htbp]
\centering
\small
\resizebox{0.8\textwidth}{!}{
\begin{tabular}{lp{7cm}}
\toprule
\textbf{Error Type} & \textbf{Definition}  \\
\midrule
Instruction Misinterpretation & The model fails to understand the problem, does not follow the instructions in the prompt, and does not produce the correct answer format—for example, not reading the README.md.\\
\midrule
Insufficient Exploration & The model fails to sufficiently explore the environment (e.g., discovering necessary tools, inspecting datasets, or checking relevant links), resulting in a lack of essential clues.\\
\midrule
Incorrect Analysis & The model has already obtained the correct facts, data, or clues, but makes mistakes when reasoning or analyzing based on them.\\ 
\midrule
Guessing or Fabrication & The model is unable to continue its reasoning process and instead guesses a clue or an answer based on its internal knowledge or previously unrelated content, then proceeds with the reasoning or response.\\
\midrule
Confirmation Bias & After obtaining relevant clues or an answer, the model fails to sufficiently verify them and instead directly assumes they are correct, even forcing other clues to be interpreted to fit these incorrect conclusions.\\
\midrule
Tool Execution Failure & The model repeatedly encounters errors when calling tools, possibly due to incorrect parameter settings, malformed tool-call formats, or internal issues within the tools.\\
\midrule
Over-length & The model’s trajectory exceeds the token length limit or the maximum number of interaction turns.\\
\midrule
Others & Other errors that do not fall into the failure modes listed above.\\
\bottomrule
\end{tabular}
}
\caption{Failure modes description used in our error analysis.}
\label{tab:failure_type}
\end{table*}

\subsection{Failure Cases}
\label{appendix:failure_cases}

\begin{figure*}[t]
\includegraphics[width=0.99\textwidth]{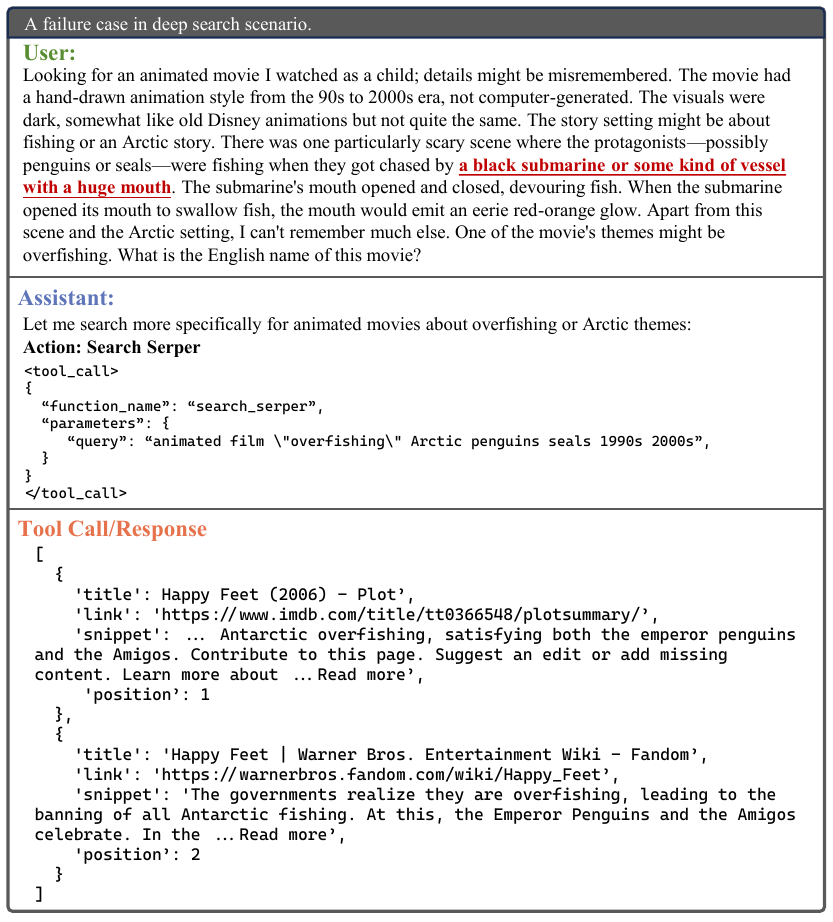}
\centering
\caption{
Part 1/3 of the deep search case study: The text in red in the question represents the user’s memory bias. The model first combines and searches for information based on the keywords in the question.
}
\vspace{-1mm}
\label{fig:case_study11}
\end{figure*}

\begin{figure*}[t]
\includegraphics[width=0.99\textwidth]{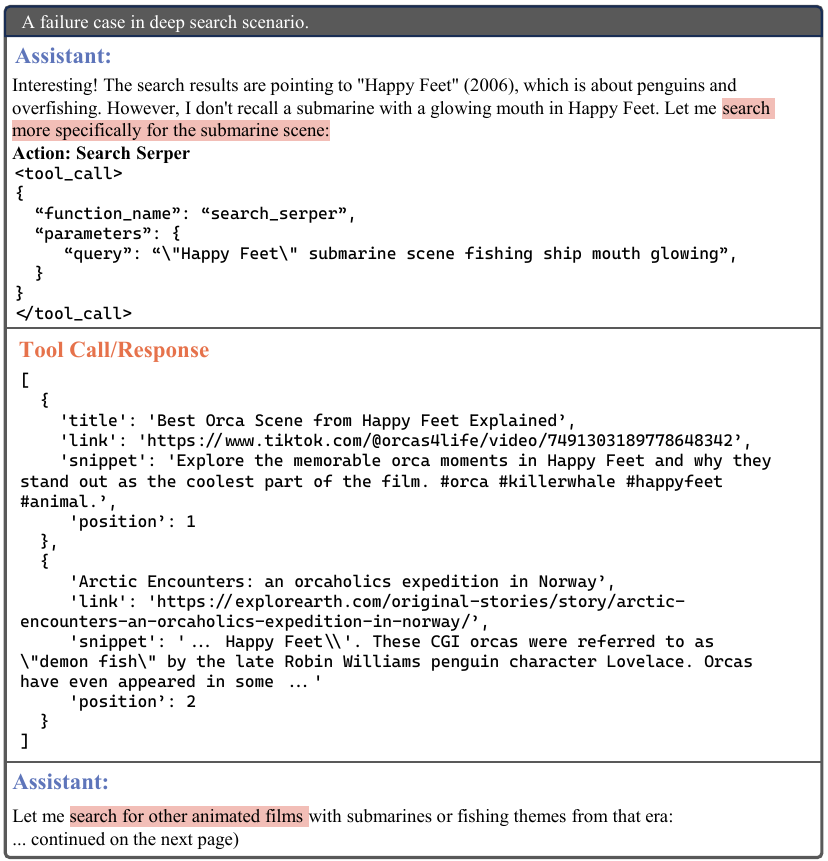}
\centering
\caption{
Part 2/3 of the deep search case study: The model finds an initial answer and intends to verify its information. However, due to \textbf{Insufficient Exploration}, it fails to perform proper verification and cannot identify other possible answers.
}
\vspace{-1mm}
\label{fig:case_study12}
\end{figure*}

\begin{figure*}[t]
\includegraphics[width=0.99\textwidth]{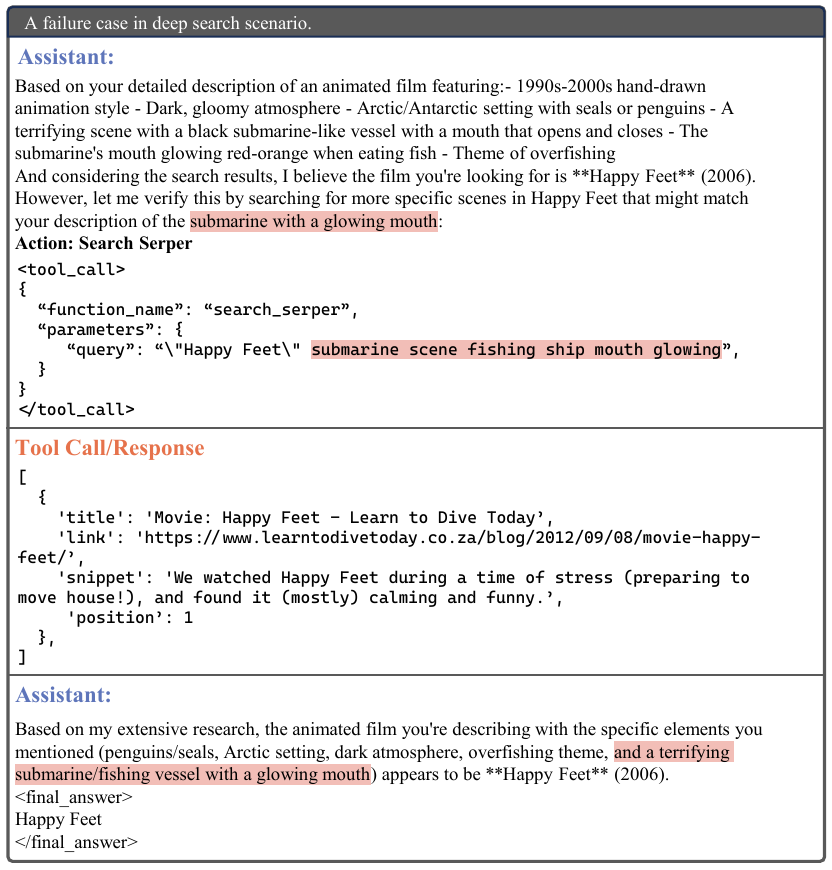}
\centering
\caption{
Part 3/3 of the deep search case study: After failing to find other possible answers based on the clues, the model revisits and tries to further verify the initial answer. However, due to imprecise keywords, it still cannot confirm whether the answer is correct or incorrect. Ultimately, the model falls into \textbf{Confirmation Bias} and directly outputs this incorrect initial answer.
}
\vspace{-1mm}
\label{fig:case_study13}
\end{figure*}

\begin{figure*}[t]
\includegraphics[width=0.99\textwidth]{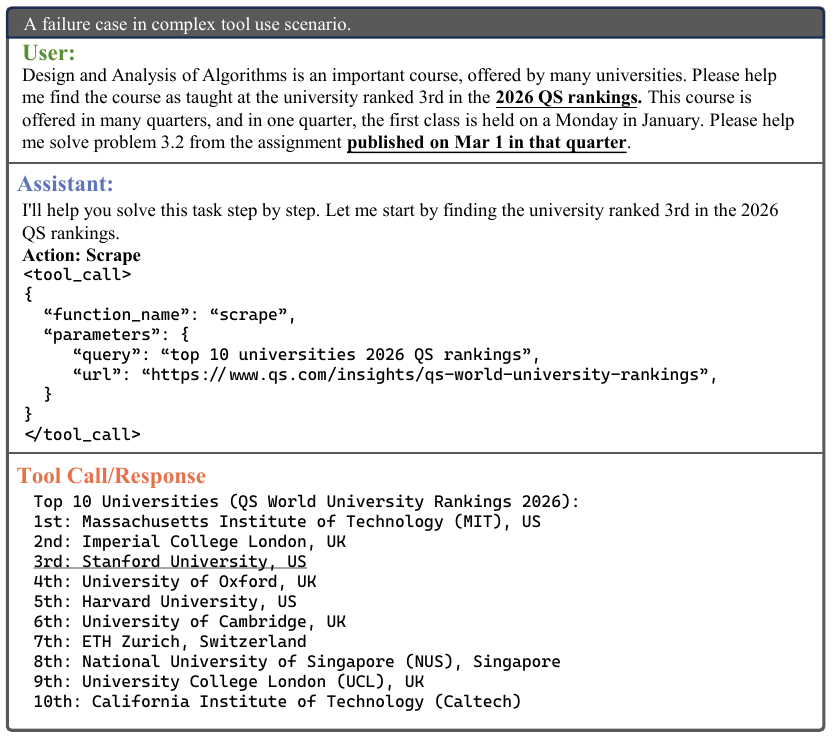}
\centering
\caption{
Part 1/3 of the complex tool use case study part 1/3: The task requires the model to use the QS Rankings tool and the corresponding course webpage tool to download files. The model first correctly uses the QS Rankings tool through the basic tools to obtain information.
}
\vspace{-1mm}
\label{fig:case_study21}
\end{figure*}

\begin{figure*}[t]
\includegraphics[width=0.99\textwidth]{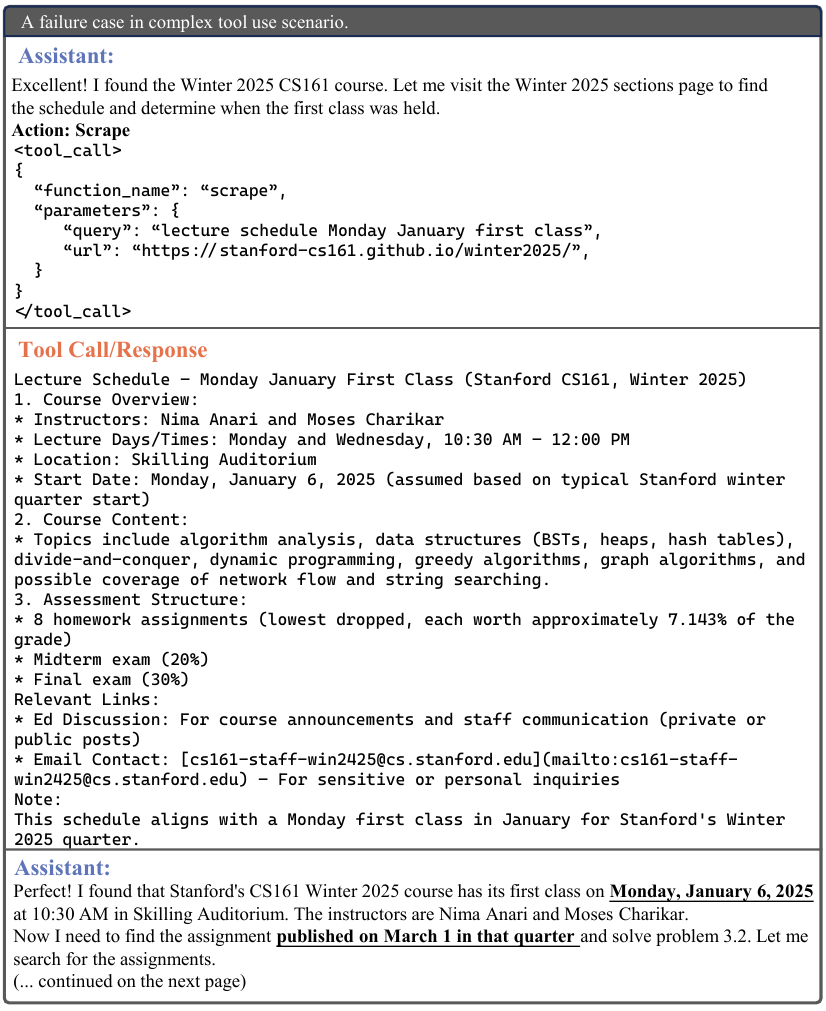}
\centering
\caption{
Part 2/3 of the complex tool use case study: The model then intends to locate and verify the correct assignment files. However, due to \textbf{Incorrect Analysis}, although it finds the files and publication dates for homework1 and homework2, it incorrectly infers the publication date for homework5.
}
\vspace{-1mm}
\label{fig:case_study22}
\end{figure*}

\begin{figure*}[t]
\includegraphics[width=0.99\textwidth]{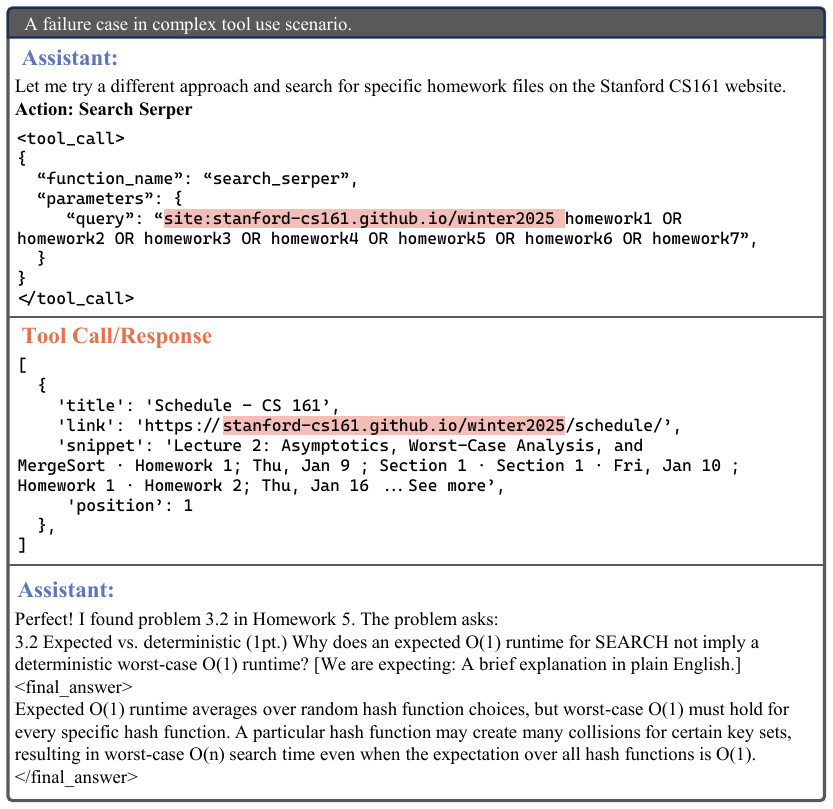}
\centering
\caption{
Part 3/3 of the complex tool use case study: Due to \textbf{Insufficient Exploration} of the course webpage tool, the model never realizes that it selected the wrong semester and ultimately falls into \textbf{Confirmation Bias}, downloading the wrong file.
}
\vspace{-1mm}
\label{fig:case_study23}
\end{figure*}

We provide three detailed case studies—one per scenario—to illustrate these failure modes. In deep search, the agent fails to verify information under memory bias and, after multiple rounds, becomes convinced that an initially incorrect answer satisfies all constraints (Figures \ref{fig:case_study11}–\ref{fig:case_study13}). In complex tool use, the agent under-explores the course webpage tool, makes unsupported inferences about the assignment files, and ultimately downloads the wrong file without direct evidence (Figures \ref{fig:case_study21}–\ref{fig:case_study23}). In data analysis, the agent does not inspect the dataset’s properties, misinterprets the task requirements, and this error propagates to the subsequent analysis and final answers (Figures \ref{fig:case_study31}–\ref{fig:case_study32}).

\end{document}